\documentclass[preprint,review,12pt]{elsarticle}
\usepackage{amsmath}
\usepackage{graphicx}
\usepackage{subfigure}
\usepackage{multirow}
\usepackage{array}
\usepackage{amssymb}
\usepackage[pagebackref=true,breaklinks=true,citecolor=blue,colorlinks,bookmarks=false]{hyperref}

\usepackage[linesnumbered,ruled,vlined]{algorithm2e}%[ruled,vlined]{
\usepackage{algpseudocode}

\graphicspath{{figures/},{figures/}}

\journal{Information Sciences}

\begin{document}

\begin{frontmatter}

\title{Action Recognition for Depth Video using Multi-view Dynamic Images}

\author{Yang~Xiao}
\ead{Yang$\_$Xiao@hust.edu.cn}
\author{Jun~Chen}
\ead{chenjun2015@hust.edu.cn}
\author{Yancheng~Wang}
\ead{yancheng$\_$wang@hust.edu.cn}
\author{Zhiguo Cao\corref{cor1}}
\ead{zgcao@hust.edu.cn}
\cortext[cor1]{Corresponding author. Tel.: +86 - 27 - 87558918}
\address{National Key Laboratory of Science and Technology on Multi-spectral Information Processing, School of Automation, Huazhong University of Science and Technology, Wuhan, Hubei, 430074, P.R. China}

\author{Joey Tianyi~Zhou\corref{}}
\ead{zhouty@ihpc.a-star.edu.sg}
\address{Institute of High Performance Computing, A*STAR, Singapore}

\author{Xiang~Bai\corref{}}
\ead{xbai@hust.edu.cn}
\address{School of Electronic Information and Communications, Huazhong University of Science and Technology, Wuhan, Hubei, 430074, P.R. China}

\begin{abstract}
Dynamic imaging is a recently proposed action description paradigm for simultaneously capturing motion and temporal evolution information, particularly in the context of deep convolutional neural networks (CNNs). Compared with optical flow for motion characterization, dynamic imaging exhibits superior efficiency and compactness. Inspired by the success of dynamic imaging in RGB video, this study extends it to the depth domain. To better exploit three-dimensional (3D) characteristics, multi-view dynamic images are proposed. In particular, the raw depth video is densely projected with respect to different virtual imaging viewpoints by rotating the virtual camera within the 3D space. Subsequently, dynamic images are extracted from the obtained multi-view depth videos and multi-view dynamic images are thus constructed from these images. Accordingly, more view-tolerant visual cues can be involved. A novel CNN model is then proposed to perform feature learning on multi-view dynamic images. Particularly, the dynamic images from different views share the same convolutional layers but correspond to different fully connected layers. This is aimed at enhancing the tuning effectiveness on shallow convolutional layers by alleviating the gradient vanishing problem. Moreover, as the spatial occurrence variation of the actions may impair the CNN, an action proposal approach is also put forth. In experiments, the proposed approach can achieve state-of-the-art performance on three challenging datasets.
\end{abstract}


\end{frontmatter}

%% \linenumbers

%% main text
\section{Introduction}
With the recently emerged low-cost depth sensors (e.g., Microsoft Kinect), action recognition using depth has attracted increasing attention in, e.g., video surveillance, multimedia data analysis, and human-machine interaction. Compared with its RGB counterpart, depth video can provide richer three-dimensional (3D) descriptive information for action characterization. Several studies~\cite{wang2014learning,veeriah2015differential,oreifej2013hon4d,yang2012recognizing,wang2016action,yang2014super} have been concerned with the advantage of 3D visual cues for action recognition from different theoretical perspectives. In practical applications, a major trend is to address the depth-based action recognition problem under \emph{high intra-class and camera view variation conditions, with a large number of action classes}. Accordingly, some challenging datasets (e.g., NTU RGB-D~\cite{Shahroudy_2016_CVPR}) have recently been proposed. Unfortunately, the performance of existing approaches on these datasets is unsatisfactory. To improve performance, one may seek a more discriminative depth-based action representation paradigm.

To better reveal action characteristics, a key issue is to adequately capture dynamic motion information. To this end, most state-of-the-art RGB-based action recognition approaches~\cite{simonyan2014two} resort to extracting dense optical flow fields~\cite{brox2011large}, and for depth video, scene flow~\cite{basha2013multi} is required for obtaining 3D motion characteristics. Nevertheless, accurate scene flow estimation is still a challenging task involving high computational cost~\cite{basha2013multi}; thus, it is not feasible for practical applications. An alternative method for resolving this is to capture the articulated 3D movement of human body skeleton joints~\cite{xia2012view,wang2014learning,vemulapalli2014human}. However, this point-based sparse description scheme may lead to information loss. Moreover, the existing human body skeleton extraction methods are still sensitive to pose and imaging viewpoint variation~\cite{oreifej2013hon4d} and may fail in certain cases. Accordingly, a more concrete motion characterization approach for depth video is required.

Dynamic imaging~\cite{bilen2016dynamic,bilen2016action} is the most recent compact video representation paradigm, based on temporal rank pooling~\cite{fernando2017rank}. It compresses a video or video clip into a single image and simultaneously maintains rich motion information. In the context of deep learning technology (i.e., deep convolutional neural networks (CNNs)), considerable success has been achieved in RGB-based action recognition~\cite{bilen2016dynamic,bilen2016action,fernando2016discriminative} at reasonable computational cost. Compared with optical flow for motion characterization, dynamic imaging has the advantages of computational efficiency and compactness. In particular, achieving higher running efficiency is relatively easy by solving linear ranking support vector machines (SVMs), and to quickly obtain approximate solutions~\cite{bilen2016action}. Furthermore, using dynamic imaging, stacking of the optical flow field sequence (as the input of the CNN for action description) may be avoided~\cite{simonyan2014two}, thus reducing the complexity of the CNN. Inspired by this, the present study extends dynamic imaging to the depth domain. To the authors' knowledge, this has not been well studied. A naive approach is to directly apply dynamic imaging to depth video as in the RGB case. However, this cannot fully exploit 3D motion properties. Accordingly, it is proposed that the raw depth video be densely projected with respect to different virtual imaging viewpoints by rotating a virtual camera around specific instances within the 3D observation space. This procedure is derived from the intrinsic 3D imaging characteristics of depth data, which cannot be achieved by RGB data. Subsequently, dynamic images are extracted from the obtained multi-view depth videos, and thus multi-view dynamic images are constructed for action characterization.

%to share the same convolutional and fully connected layers.

The high adaptability to CNNs is another major advantage of dynamic images, which ensures strong discriminative power. In~\cite{bilen2016dynamic,bilen2016action,fernando2016discriminative}, the dynamic images from RGB channels are fed into a one-stream CNN model~\cite{Jia2014Caffe} for action representation. In the present study, it is argued that this paradigm is not optimal for the proposed multi-view dynamic imaging. Deep insight is obtained from the CNN learning perspective, and the gradient vanishing problem~\cite{bengio1994learning,glorot2010understanding} in the training is the critical issue faced by the deep neural network, particularly when the training sample size is not sufficient. Therefore, the shallow layers (i.e., the convolutional layers) in the CNN may not be adequately tuned~\cite{he2016deep}, thus impairing its visual pattern capture capacity.  Regarding the aim of the present study, the scale of the existing depth action datasets~\cite{Shahroudy_2016_CVPR,rahmani2016histogram,wang2014cross} (i.e., 56000 samples at most) is still relatively small compared with the large-scale image recognition datasets (e.g., Imagenet with millions of samples) suitable for CNN training. Moreover, compared with the RGB dynamic images in~\cite{bilen2016dynamic,bilen2016action,fernando2016discriminative}, multi-view dynamic images in depth video are of higher complementarity. Hence, the motivation is to better exploit this complementarity property and improve the training effectiveness on the convolutional layers in the case that the training samples are not abundant. To this end, a novel CNN learning model is proposed. In particular, the dynamic images from different imaging viewpoints share the same convolutional layers but correspond to different fully connected layers. During training, the fully connected layers of different viewpoints will be iteratively tuned, whereas the shared convolutional layers are consistently tuned during the entire training phase, and the training error from different viewpoints can be back-propagated to the convolutional layers with more chances. Thus, the gradient vanishing problem can be alleviated to some degree. Consequently, the output of the fully connected layers will be employed as the visual feature for action recognition, being input into the SVM with principal component analysis (PCA).

\begin{figure*}[t]
	%h 此处 t 页定 b 页低 p独立一页
	\centering
	\includegraphics[width=13.3cm]{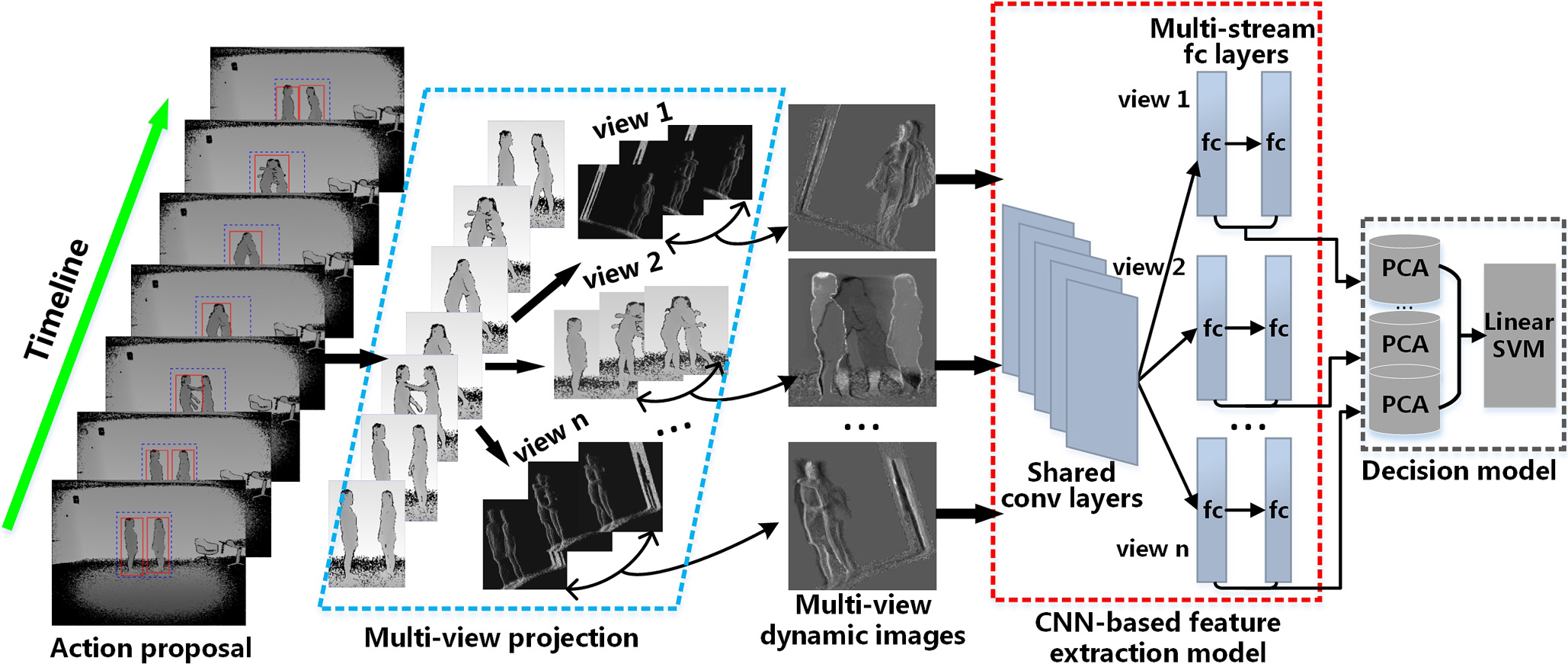}
	
	\caption{Technical pipeline diagram of the proposed action recognition method for depth video using multi-view dynamic images.}
	\label{figure10}
\end{figure*}

Generally, actions take place at different spatial locations in varying scene conditions. Dynamic imaging can fade the background effect but is still spatial-sensitive. This is the case in CNNs as well~\cite{cimpoi2015deep}. To counter the effect of spatial variation, an action proposal method is put forth in this study. Specifically, an off-the-shelf object detector faster R-CNN~\cite{ren2015faster} is first used to detect humans per frame, owing to its generality and robustness. Subsequently, the resulting human detection bounding boxes are spatial-temporally merged to construct the action proposal volume. It is noteworthy that the human--human and human--object interaction information can still be maintained within the action proposals. The dynamic image is subsequently extracted from the action proposal volume (not from the entire video) and is fed to the CNN. The main technical pipeline diagram of the  proposed scheme is shown in Fig.~\ref{figure10}.

The proposed action recognition approach based on multi-view dynamic images was tested on three challenging depth datasets (i.e., NTU RGB-D~\cite{Shahroudy_2016_CVPR}, Northwestern--UCLA~\cite{wang2014cross}, and UWA3DII~\cite{rahmani2016histogram}). The experimental results demonstrate that the method can achieve state-of-the-art performance on all datasets. The effectiveness of the method is also investigated by an ablation study.

The main contributions of this study are:
\begin{itemize}
\item Dynamic images are extended to the depth domain for action recognition. Multi-view dynamic imaging is proposed for obtaining 3D motion characteristics for action description.
\item A novel CNN learning model is proposed to enhance the training effectiveness on multi-view dynamic images;
\item An action proposal approach is put forth to counter the effect of spatial variation.
\end{itemize}

The source code and supporting material for this paper can be accessed at \url{https://github.com/3huo/MVDI}.

The rest of this paper is organized as follows. Sec.~\ref{sec:relatedwork} introduces related work. Sec.~\ref{sec:mvdi} illustrates the concept of multi-view dynamic image in detail. The proposed CNN learning model for multi-view dynamic images is presented in Sec.~\ref{sec:cnnmodel}. The action proposal approach is introduced in Sec.~\ref{sec:actionproposal}. Experiments and discussion are presented in Sec.~\ref{sec:experiments}. Sec.~\ref{sec:conclusion} concludes the paper.

\section{Related Work} \label{sec:relatedwork}

Depth-based action recognition approaches can be generally categorized into three main groups: skeleton-based, raw depth-video-based, and their combination.

\textbf{Skeleton-based.} Under this paradigm, 3D human body skeleton joints are first extracted from the depth frames for action characterization. Using the skeleton information, Wang~\emph{et al.}~\cite{wang2014learning} extracted the spatial-temporal pairwise distances between the skeleton joints and mined the most discriminative joint combinations for specific action classes. Vemulapalli~\emph{et al.}~\cite{vemulapalli2014human} and Huang~\emph{et al.}~\cite{huang2016deep} proposed extracting the discriminative Lie group action representation from the manifold learning perspective. Weng~\emph{et al.}~\cite{wengspatio} resorted to nearest neighbor search to categorize actions. A recent research trend is to capture the spatial-temporal evolution of skeleton joints using recurrent neural networks (RNNs)~\cite{veeriah2015differential} and long--short-term memory networks (LSTM)~\cite{Shahroudy_2016_CVPR}.  Despite its noticeable progress, the skeleton-based approach still suffers from information loss and potential skeleton joint extraction failure. Moreover, the CNN cannot benefit from this paradigm and improve its performance.

\textbf{Raw depth-video-based.} Within this group, spatial-temporal features for action description are captured from the raw depth video directly without extracting the skeletons. Oreifej~\emph{et al.}~\cite{oreifej2013hon4d} proposed HON4D for action characterization by computing the histogram of oriented normal vectors in the 4D space (i.e., XYZ-T). Later, Rahmani~\emph{et al.}~\cite{rahmani2014hopc} refined HON4D by encoding the histogram of oriented principal components (HOPC) within the 3D volume around each cloud point. To counter the effect of imaging noise, camera view variation, and cluttered background, Lu~\emph{et al.}~\cite{lu2014range} proposed a binary descriptor by performing pairwise comparisons among the 3D cloud points. In~\cite{yang2012recognizing}, the concept of depth motion map (DMM) is proposed to compress the depth video into a single image by intuitively aggregating the video frames using sum pooling. The histogram of oriented gradients (HOG) descriptor is consequently extracted on the DMM from three orthogonal projection planes. In fact, DMMs are in a sense similar to dynamic images. However, it does not involve temporal evolution characteristics. HOG was replaced with the FV-encoded LBP in~\cite{chen2015action}. Wang~\emph{et al.} Recently, in ~\cite{wang2015convnets,wang2016action} CNNs were applied to DMMs, as in the present study. The main differences are as follows. First, the superiority of dynamic images over DMMs is verified for action characterization in depth video. Second, a novel CNN learning model that better handles multiple-view dynamic images is constructed. Finally, the action proposal procedure is not involved in~\cite{wang2015convnets,wang2016action}.

\textbf{Skeleton and raw depth video fusion manner.} To better use the information from the skeletons and raw depth video, Wang~\emph{et al.}~\cite{wang2012robust} proposed extracting the local occupancy pattern  from the 3D volume space along the skeleton joints. Following this, Yang~\emph{et al.}~\cite{yang2014super} chose to extract the super normal vector (SNV) descriptor along the skeleton joints. Althloothi~\emph{et al.}~\cite{althloothi2014human} and Chaaraoui~\emph{et al.}~\cite{chaaraoui2013fusion} proposed fusing the skeleton and silhouette shape features.

The present study falls into the second group. Skeleton extraction is not required; thus, robustness is ensured, and richer representative depth video information is involved. To enhance the discriminative power of spatial-temporal features, dynamic imaging and a CNN model are employed. For a more complete survey on action recognition using depth data, readers are referred to~\cite{wang2018rgb-d-based}.

Obtaining multi-view dynamic images from depth video can be regarded as the most important contribution of this study, as this allows the action recognition task to be considered within the multi-view learning framework, thus enhancing performance through late feature fusion. Multi-view learning can also be applied to numerous other visual recognition problems. For instance, Yu~\emph{et al.}~\cite{yu2014click,yu2017deep} proposed novel sparse coding and deep metric learning approaches to fuse multimodal information and facilitate image ranking. Moreover, a deep autoencoder method~\cite{hong2015multimodal} was also proposed to integrate multimodal cues and map two-dimensional (2D) images to 3D poses. Compared with these methods, the focus of the present study is on alleviating the gradient vanishing problem in the CNN by multi-view learning.

\section{Multi-View Dynamic Images} \label{sec:mvdi}

To concretely capture the motion information in depth video, the concept of dynamic image is extended from the RGB domain to the depth domain. By densely projecting the depth video with respect to multiple virtual imaging points, multi-view dynamic images are extracted involving more discriminative information. Moreover, the training sample size can be increased for better CNN tuning.

\subsection{Dynamic images}
To use CNNs for action characterization, Bilen~\emph{et al.}~\cite{bilen2016dynamic,bilen2016action} proposed the concept of dynamic images to capture spatial-temporal dynamic evolution information in a video. A major advantage of dynamic images is that they can summarize a video or video clip into a single static image. Intuitively, this can improve the efficiency of the CNN for both the off-line training and the online test phases. Let $I_i,\cdot\cdot\cdot,I_T$ denote the video frames, and $V_t=\frac{1}{t}\times\sum_{i=1}^{t}I_i$ be the frame average until time $t$. Subsequently, a video ranking score function at each time $t$ is defined as
\begin{equation}
\label{eq:ranking_score}
S(t|\textbf{u})=\left\langle \textbf{u}, V_t\right\rangle,
\end{equation}
where $\textbf{u}\in\mathbb{R}^d$ is the ranking parameter vector. $\textbf{u}$ is learned from the specific video to reflect the ranking relationship among the video frames. The criterion is that the later frames are associated with the larger ranking scores as
\begin{equation}
\label{eq:ranking_relationship}
q>t \Rightarrow S(q|\textbf{u})>S(t|\textbf{u}).
\end{equation}
The learning procedure of $\textbf{u}$ is subsequently formulated as a convex optimization problem based on the framework of RankSVM as
\begin{equation}
\begin{aligned}
\textbf{u}^*&=\mathop{\text{argmin}}\limits_\textbf{u} \frac{\lambda}{2}\parallel \textbf{u} \parallel^2+\\ &{\frac{2}{T(T-2)}\times \sum_{q>t} \text{max}\left\{0,1-S(q|\textbf{u})+S(t|\textbf{u})\right\}}.
\end{aligned}
\label{eq:rankSVM}
\end{equation}

\begin{figure}[t]
\centering  % by Jianxin -- never use \begin{center}\end{center}; use \centering
\subfigure[Hand waving] {\label{fig:di_example(a)}
\includegraphics[width=4.4cm]{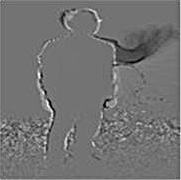}}
\subfigure[Wearing shoe] {\label{fig:di_example(b)}
\includegraphics[width=4.4cm]{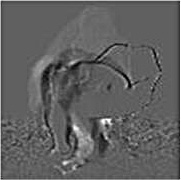}}
\subfigure[Kicking other person] {\label{fig:di_example(c)}
\includegraphics[width=4.4cm]{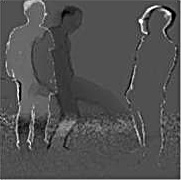}}
\subfigure[Hugging other person] {\label{fig:di_examplep(d)}
\includegraphics[width=4.4cm]{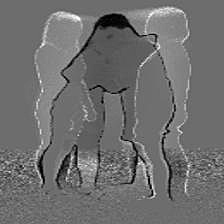}}
\caption{Dynamic image samples of four actions in depth video.}
\label{fig:di_example}
\end{figure}

In particular, the first term is the regularizer usually employed in SVMs and the second term is the hinge-loss soft-counting of the number of pairs $q>t$ incorrectly ranked by the scoring function, i.e., those that do not satisfy $S(q|\textbf{u})>S(t|\textbf{u})+1$. Optimizing Eqn.~\ref{eq:rankSVM} can map the video frames $I_i,\cdot\cdot\cdot,I_T$ to a single vector $\textbf{u}^*$. In fact, $\textbf{u}^*$ encodes the dynamic evolution information from all frames. Spatially reordering $\textbf{u}^*$ from 1D to 2D yields dynamic images for video representation. Dynamic images have already demonstrated their superiority for action characterization in RGB video ~\cite{bilen2016dynamic,bilen2016action,fernando2016discriminative}. In the present study, it is demonstrated that this can also be extended to depth video. Fig.~\ref{fig:di_example} shows the dynamic image samples of four actions (i.e., \emph{ ``hand waving", ``wearing shoe", ``kicking other person"}, and \emph{``hugging other person"}) in the depth videos from the NTU RGB-D dataset~\cite{Shahroudy_2016_CVPR}. It can be observed that the discriminative dynamic motion information in the video frames can be obtained from a single dynamic image, and simultaneously the background is suppressed. Moreover, the motion temporal order is also reflected by the gray-scale value.

However, intuitively applying dynamic imaging to the depth domain cannot fully exploit the 3D visual clues contained in depth video. To address this, it is proposed that the depth video be densely projected with respect to multiple virtual imaging viewpoints in the 3D observation space. Dynamic images are then extracted from the obtained depth videos, and thus multi-view dynamic images are constructed.

\begin{figure}[t]
	%h 此处 t 页定 b 页低 p独立一页
	\centering
	\includegraphics[width=7cm]{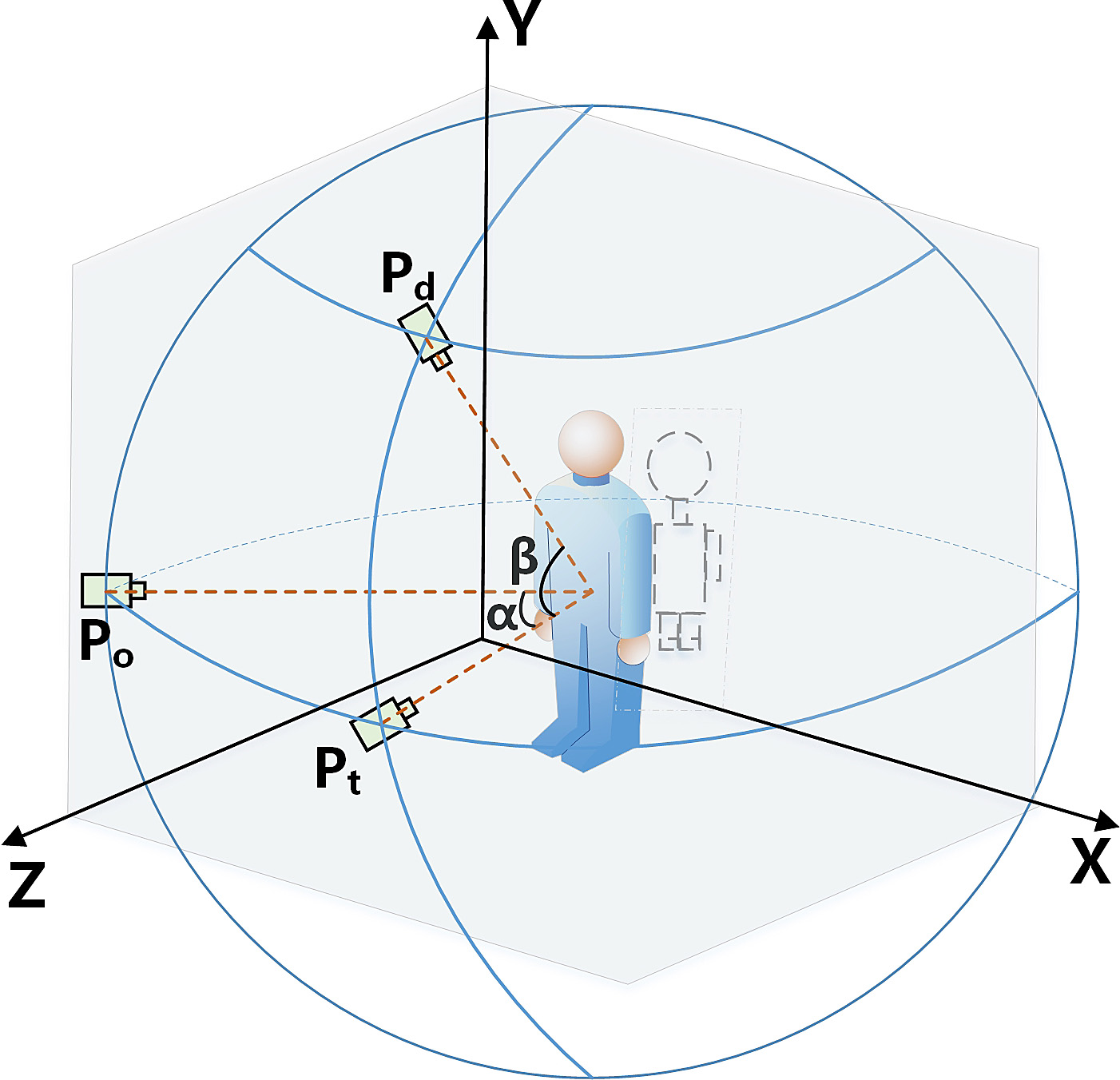}
	
	\caption{Rotation of the virtual camera within 3D space to mimic different imaging viewpoints in depth video.}
	\label{fig:camera_rotate}
\end{figure}

\subsection{Multi-view projection on depth video} \label{sec:mvp_dv}

Unlike RGB video, depth video can be observed from different viewpoints. This can be achieved by rotating the virtual camera around  specific instances in 3D space, as shown in Fig.~\ref{fig:camera_rotate}. This is equivalent to rotating the 3D cloud points within the depth frames. Consequently, a series of synthesized depth videos can be generated by multi-view projection in the raw video. This facilitates 3D discriminative visual information mining as well as data augmentation for CNN training.

\begin{figure}[t]
	\centering

    \subfigure[$\alpha=0^{\circ};\beta=0^{\circ}$]{\includegraphics[width=3.5cm]{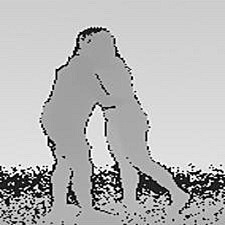}}
    \subfigure[$\alpha=0^{\circ};\beta=5^{\circ}$]{\includegraphics[width=3.5cm]{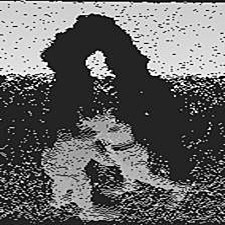}}
    \subfigure[$\alpha=-10^{\circ};\beta=0^{\circ}$]{\includegraphics[width=3.5cm]{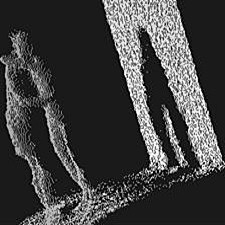}}
    \subfigure[$\alpha=-10^{\circ};\beta=5^{\circ}$]{\includegraphics[width=3.5cm]{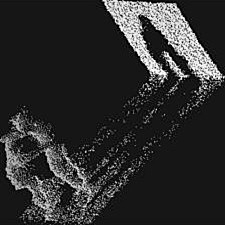}}
    \subfigure[$\alpha=10^{\circ};\beta=0^{\circ}$]{\includegraphics[width=3.5cm]{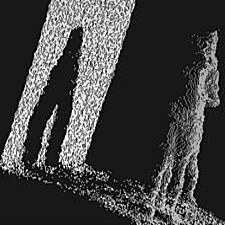}}
	\subfigure[$\alpha=10^{\circ};\beta=5^{\circ}$]{\includegraphics[width=3.5cm]{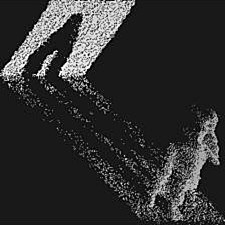}}
	
	\caption{Multi-view projection results on depth frame. In particular, ``$\alpha=0^{\circ};\beta=0^{\circ}$" corresponds to the raw depth frame.}
	
\label{fig:mvp_result}
\end{figure}

As shown in Fig.~\ref{fig:camera_rotate}, if the virtual camera is to be moved from $P_o$ to $P_d$, $P_o$ can be first moved to $P_t$ with a rotation angle $\alpha$ around the $\textbf{Y}$ axis, and then $P_t$ can be moved to $P_d$ with a rotation angle $\beta$ around the $\textbf{X}$ axis. The corresponding rotation matrices for the 3D point coordinate transformation are given by

\begin{equation}
\label{eq:rotation_x}
\textbf{R}_{\textbf{x}}=\begin{bmatrix}
\cos(\beta)&0&\sin(\beta)\\0&1&0\\-\sin(\beta)&0&\cos(\beta)
\end{bmatrix},
\end{equation}
and
\begin{equation}
\label{eq:rotation_y}
\textbf{R}_{\textbf{y}}=\begin{bmatrix}
1&0&0\\0&\cos(\alpha)&-\sin(\alpha)\\0&\sin(\alpha)&\cos(\alpha)
\end{bmatrix},
\end{equation}
where the right-handed coordinate system is used for rotation, and the original camera viewpoint is regarded as the rotation angle origin. Consequently, the coordinates of a 3D point at $\left(P_x,P_y,P_z\right)^T$ after viewpoint rotation are transformed as
\begin{equation}
\left(\overrightarrow{P_x},\overrightarrow{P_y},\overrightarrow{P_z}\right)^T
=\textbf{R}_{\textbf{x}}\textbf{R}_{\textbf{y}}\left(P_x,P_y,P_z\right)^T,
\end{equation}
where $\overrightarrow{P_x}$, $\overrightarrow{P_y}$, and $\overrightarrow{P_z}$ can be regarded as the new screen coordinates, and the corresponding depth value, respectively, for the synthesized depth frames. The proposed multi-view projection approach for depth video is similar to  that in~\cite{wang2015convnets,wang2016action}. However, it is noteworthy that it does not transform $\left(P_x,P_y,P_z\right)^T$ to real-world coordinates as in~\cite{wang2015convnets,wang2016action} because this requires the focal length of the depth camera, which is not always available in practical applications. Thus, even though the proposed method is relatively easier and more convenient in applications, it still has high performance. Fig.~\ref{fig:mvp_result} shows some multi-view projection results in a specific depth frame corresponding to different $\alpha\text{-}\beta$ combinations. In particular, ``$\alpha=0;\beta=0$" corresponds to the raw depth frame. It can be observed that more representative visual information is involved in the ensemble of the synthesized depth frames. Moreover, the sample size for a specific action can be significantly increased, which facilitates the alleviation of the data hungriness problem during CNN training, as previously mentioned.

\begin{figure}[t]
	\centering

    \subfigure[$\alpha=0^{\circ};\beta=0^{\circ}$]{\includegraphics[width=3.5cm]{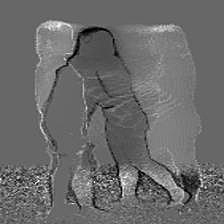}}
    \subfigure[$\alpha=0^{\circ};\beta=5^{\circ}$]{\includegraphics[width=3.5cm]{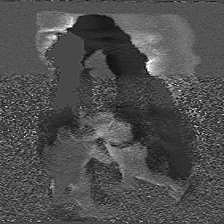}}
    \subfigure[$\alpha=-10^{\circ};\beta=0^{\circ}$]{\includegraphics[width=3.5cm]{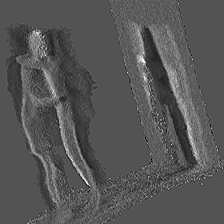}}
    \subfigure[$\alpha=-10^{\circ};\beta=5^{\circ}$]{\includegraphics[width=3.5cm]{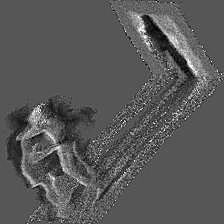}}
    \subfigure[$\alpha=10^{\circ};\beta=0^{\circ}$]{\includegraphics[width=3.5cm]{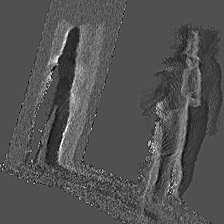}}
	\subfigure[$\alpha=10^{\circ};\beta=5^{\circ}$]{\includegraphics[width=3.5cm]{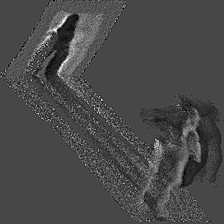}}
	
	\caption{Dynamic images corresponding to the multi-view projection results in Fig.~\ref{fig:mvp_result}.}
	
\label{fig:mvdi_result}
\end{figure}

\subsection{Multi-view dynamic image extraction} \label{sec:mvdi_extraction}

\begin{figure}[t]
	%h 此处 t 页定 b 页低 p独立一页
	\centering
	\includegraphics[width=9cm]{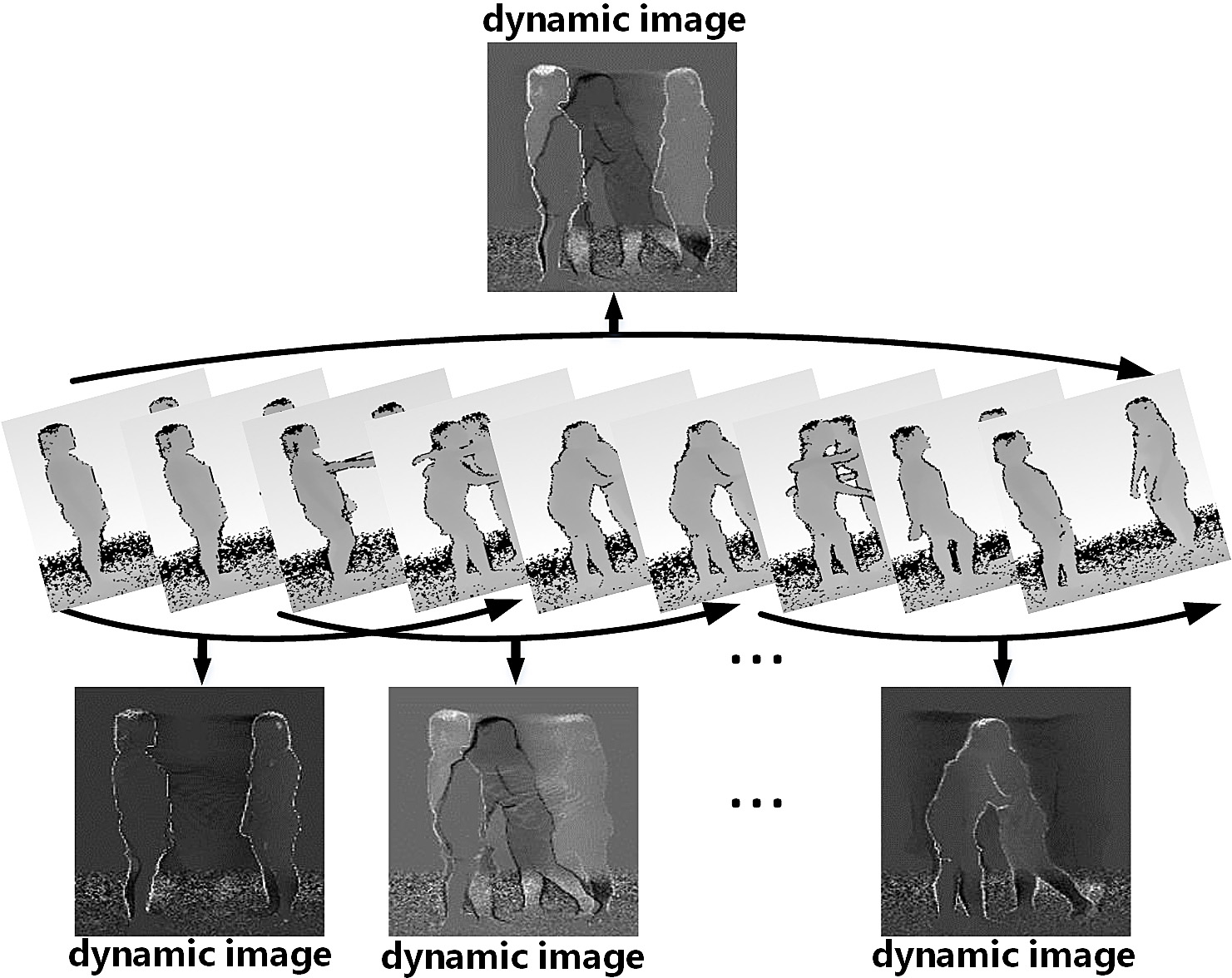}
	\caption{Temporal split for multi-view dynamic image extraction.}
	\label{fig:temporal_split}
\end{figure}

After the multi-view projection procedure in the depth video has been completed, dynamic images are individually extracted from the synthesized multi-view depth videos (including the raw video), as shown in Fig.~\ref{fig:mvdi_result}. The ensemble of the obtained single-view dynamic images is termed as multi-view dynamic images for action characterization. Moreover, to involve richer temporal representative information, the single-view videos are split into overlapping temporal segments. The dynamic images are simultaneously extracted from the temporal segments and the entire single-view video, as shown in Fig.~\ref{fig:temporal_split}. In particular, each video is empirically split into four temporal segments with an overlap ratio of 0.5. Following~\cite{bilen2016dynamic,bilen2016action}, the temporal segments share the same action category label as the raw single-view video. In fact, the temporal split procedure is also used for data augmentation for CNN training to improve performance.

\section{Multi-view Dynamic Image Adaptive CNN Learning Model} \label{sec:cnnmodel}

After the multi-view dynamic images have been extracted, they are fed into the CNN for action characterization. Hence, an adapted CNN learning model is required to ensure performance. Two issues require attention. First, the scale of the existing depth action datasets~\cite{Shahroudy_2016_CVPR,rahmani2016histogram,wang2014cross} (i.e., 56000 samples at most) is still limited; thus, the requirements of data-hungry CNN training cannot be fully met. In fact, this may aggravate the effect of the gradient vanishing problem~\cite{bengio1994learning,glorot2010understanding}, particularly on the shallow convolutional layers that capture visual patterns. Second, even though the extracted multi-view dynamic images are of high complementarity, as shown in Fig.~\ref{fig:mvdi_result}, the conventional one-stream CNN model~\cite{Jia2014Caffe} employed in~\cite{bilen2016dynamic,bilen2016action,fernando2016discriminative} cannot capture this well. In~\cite{wang2015convnets,wang2016action}, a multi-stream CNN model is proposed to alleviate this problem on multi-view DMMs. That is, the DMMs from different viewpoints correspond to the individual convolutional and fully connected layers. However, this model still suffers from insufficient training data. Accordingly, the primary motivation for this study is to better exploit the complementary information from different viewpoints to enhance CNN training when the training samples are insufficient.

\begin{figure}[t]
	%h 此处 t 页定 b 页低 p独立一页
	\centering
	\includegraphics[width=10cm]{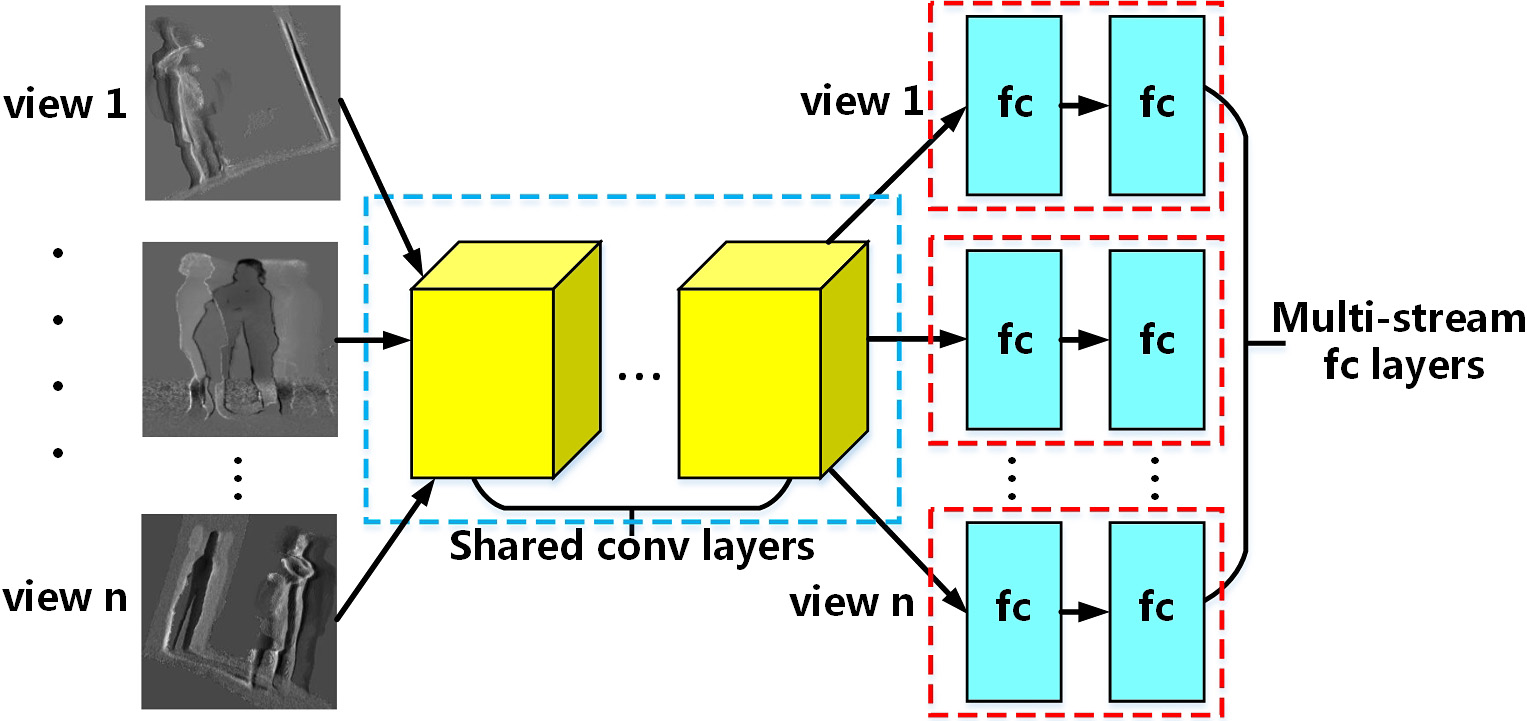}
	
	\caption{Multi-view dynamic image adaptive CNN learning model.}
	\label{fig:cnn_model}
\end{figure}

\begin{algorithm}[t]
    \footnotesize
	\caption{Training procedure of the proposed multi-view dynamic image adaptive CNN learning model}
	\label{Alg:CNN-train}
	%\KwIn{the head proposal pixel set $S=\{s_i,i\in\left\{1,2,...N\right\}\}$, the average head radius $r_h$;}
	\KwIn{dynamic image training sample sets $\left \{ DI_1,DI_2,...,DI_n\right \}$ from the $n$ virtual imaging viewpoints; $n$ sub-CNN models $\left \{ CNN_1,CNN_2,...,CNN_n\right \}$ pre-trained on Imagenet, with the same structure and parameter setting;}
	\KwOut{the tuned $n$ sub-CNN models $\left \{ CNN_1,CNN_2,...,CNN_n\right \}$ for action characterization;}
%	$I_{h} = Opening(I_{h})$ // Execute morphological opening operation on $I_{h}$\;
    \For{each training iteration} {
        \For{training sample set $DI_i$ that corresponds to the $i$-th viewpoint } {
        \If{i$=$1;}{
		$CNN_1$ inherits the convolutional layers of $CNN_n$ from the last training iteration\;
            Train $CNN_1$ using $DI_1$;
		}
        \Else
		 	{
		 		$CNN_i$ inherits the convolutional layers of $CNN_{i-1} $ within the current training iteration\;
                Train $CNN_i$ using $DI_i$;
		 	}
        }		
	}
	
	Return $\left \{ CNN_1,CNN_2,...,CNN_n\right \}$\;
\end{algorithm}

Thus, a novel multi-view dynamic image adaptive CNN learning model is proposed in Fig.~\ref{fig:cnn_model}. In particular, the dynamic images from different viewpoints share the same convolutional layers (specifically, the same convolutional filters), but correspond to different fully connected layers. The intuition is that the shared convolutional layers capture the discriminative fundamental visual patterns among the multi-view dynamic images, whereas the multi-stream fully connected layers reflect complementarity. During the training phase, the fully connected layers of different viewpoints will be iteratively tuned, whereas the shared convolutional layers are consistently trained. In particular, in this learning model, each imaging viewpoint corresponds to a sub-CNN model. All the sub-CNN models with the same structure can be trained independently in an end-to-end manner, but share the same convolutional layers. Let the sub-CNN models be denoted by $\left \{ CNN_1,CNN_2,...,CNN_n\right \}$. During each iteration of the training phase, $CNN_1$ is trained first. $CNN_2$ is subsequently trained by inheriting the acquired convolutional layers of $CNN_1$. This procedure is applied to the remaining sub-CNN models in the current training iteration. When the next training iteration starts, $CNN_1$ inherits the convolution layers of $CNN_n$ from the previous iteration. This recursive procedure continues until the entire training phase of the proposed CNN model is completed. Algorithm~\ref{Alg:CNN-train} shows the entire training procedure in detail. The proposed method can be regarded as a hybrid of the one-stream CNN model in~\cite{bilen2016dynamic,bilen2016action,fernando2016discriminative} and the multi-stream model in~\cite{wang2015convnets,wang2016action}.

Its main advantages are as follows. First, compared with the one-stream model in~\cite{bilen2016dynamic,bilen2016action,fernando2016discriminative}, the multi-stream fully connected layers in the proposed approach maintain the complementary information among the viewpoints. It is noteworthy that the output of the multi-stream fully connected layers will be concatenated as a visual feature for action recognition with PCA and SVMs in the proposed method. Furthermore, compared with the multi-stream model in~\cite{wang2015convnets,wang2016action}, in the proposed method, the training error from different viewpoints can be back-propagated to the convolutional layers with better chance. Thus, the gradient vanishing problem can be alleviated to some degree, thereby enhancing the training  effectiveness on the convolutional layers.

%\begin{algorithm}[t]
%	\caption{ Training procedure of the proposed multi-view dyynamic images adapted CNN learning model}
%	\begin{algorithmic}%[1]
%		\KwIn Five view groups datasets $\left \{ D_1,...,D_5\right \}$ and five CNN models $\left \{ M_1,...,M_5\right \}$.
%		\For{each interation}
%		\For{each view Group $D_i$ and CNN model $M_i$}
%		\State 1: Set convolutional layer parameters of $M_i$ equals to $M_{i-1}$
%		\State 2: Training $M_i$ using $D_i$
%		\EndFor
%		\EndFor
%		\Ensure Five shared Multi-view Model
%	\end{algorithmic}
%	\label{algorithm1}
%\end{algorithm}

\section{Spatial-temporal Action Proposal} \label{sec:actionproposal}

\begin{figure}[t]
	%h 此处 t 页定 b 页低 p独立一页
	\centering
	\includegraphics[width=9cm]{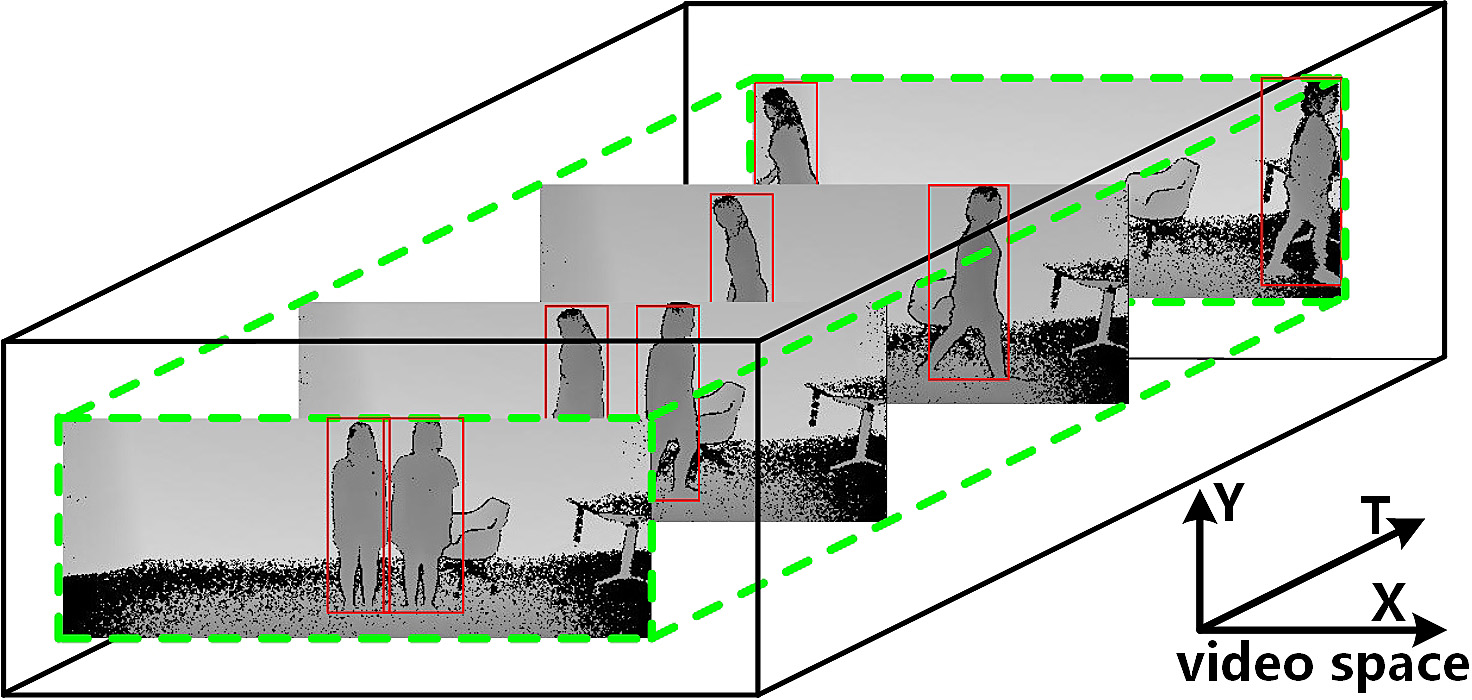}
	
	\caption{Action proposal for ``\emph{walking apart from each other}" from the NTU RGB-D~\cite{Shahroudy_2016_CVPR} dataset. In particular, the red bounding boxes are the human detection results using the faster R-CNN, and the green cubic is the action proposal region.}
	\label{fig:action_proposal}
\end{figure}

In~\cite{bilen2016dynamic,bilen2016action,fernando2016discriminative}, dynamic image extraction is directly performed in the entire video frames. However, this is not optimal for the subsequent CNN-based visual feature extraction procedure because actions of the same category may take place at different spatial locations and in varying scene conditions. Dynamic images can fade the background effect but are still spatial-sensitive. This is also the case in CNNs~\cite{cimpoi2015deep}, particularly on the shallow convolutional layers, and hinders stable action representation. Consequently, effective action proposal is required.

According to the principal idea that actions are carried out by humans, a spatial-temporal action proposal approach is set forth. In particular, human detection is first performed in each depth frame. Then, the resulting human detection bounding boxes are spatial-temporally linked and merged. That is, the minimal spatial-temporal cubic volume that closely covers all bounding boxes is considered the action proposal result, as shown in Fig.~\ref{fig:action_proposal}. Consequently, the multi-view dynamic images are extracted only from the obtained action proposal region with certain extensions\footnote{Within each frame, the four sides of the action proposal bounding box will be extended by 30 pixels evenly to involve more discriminative information.}, not form the entire depth video. Generally, the action items of human--human interaction and human--object interaction can be intrinsically involved in the proposed action proposal approach. Specifically, the off-the-shelf faster R-CNN~\cite{ren2015faster} is employed as a human detector owing to its high object detection capacity. It is noteworthy that this study primarily focuses on addressing the action recognition not spatial-temporal action detection in untrimmed video. Accordingly, all the training and test action video samples involved were trimmed in advance, with definite starting and ending time points. In particular, the action proposal procedure is performed during the entire duration of each video sample, both for training and testing. Thus, temporal action stride need not be considered.

\section{Experiments} \label{sec:experiments}

In the experiments, the focus will be on verifying the discriminative power of multi-view dynamic images for action characterization in depth video. Three challenging action datasets were used, specifically, NTU RGB-D~\cite{Shahroudy_2016_CVPR}, Northwestern--UCLA~\cite{wang2014cross}, and UWA3DII~\cite{rahmani2016histogram}. Although both RGB and depth information is involved in these datasets, only the depth visual cues are taken into consideration. If not otherwise specified, the available samples in each dataset are split into the training set and the testing set according to the original principles in~\cite{Shahroudy_2016_CVPR,wang2014cross,rahmani2016histogram}.

\begin{table}[t]
	\centering
    \linespread{1.05}\selectfont
    \footnotesize
	\caption{$\alpha$--$\beta$ view group division according to the value of $\alpha$.}
	\begin{tabular}{cc}
		\hline
		View group ($\beta=0^{\circ}$) & $\alpha$ \\
		\hline
		Group 1 & $-90^{\circ}, -40^{\circ}$ \\

		Group 2 & $ -20^{\circ}, -10^{\circ}, -5^{\circ}$ \\
	
		Group 3 & $0^{\circ}$ \\
		
		Group 4 & $5^{\circ}, 10^{\circ}, 20^{\circ}$ \\
		
		Group 5 & $40^{\circ}, 90^{\circ}$ \\
		\hline
	\end{tabular}
	\label{tab:view-group}
\end{table}

When multi-view projection is performed in depth video, $\beta$ is empirically set to $0^{\circ}$, and $\alpha$ to $-90^{\circ}$, $-40^{\circ}$, $-20^{\circ}$, $-10^{\circ}$, $-5^{\circ}$, $0^{\circ}$, $5^{\circ}$, $10^{\circ}$, $20^{\circ}$, $40^{\circ}$, and $90^{\circ}$. It is noteworthy that the tuple $\left(\alpha=0^{\circ},\beta=0^{\circ}\right)$ corresponds to the raw depth video. Furthermore, the $\alpha$--$\beta$ tuples are divided into five view groups, as listed in Table~\ref{tab:view-group}, according to the value of $\alpha$. During training, the five view groups share the same convolutional layers but correspond to individual fully connected layers, as shown in Fig.~\ref{fig:cnn_model}. The main reason for merging the adjacent $\alpha$$\beta$ tuples into groups is to restrict model size.

\begin{table}[t]
	\linespread{1.05}\selectfont
	\centering
	\scriptsize
	\caption{{\bfseries CNN-F model architecture.} It contains five convolutional layers (conv1-5) and three fully connected layers (fc6-8). The details of each convolutional filter are listed in three sub-rows: the first sub-row specifies the number of convolution filters and their receptive field size in the form of ``$num\times size\times size$"; the second sub-row indicates the convolution stride (i.e., ``st.") and the spatial padding (i.e., ``pad."); the third sub-row indicates the max-pooling down-sampling factor and whether local response normalization (LRN)~\cite{Krizhevsky2012ImageNet} is applied. fc6-8, fc6, and fc7 are regularized using dropout~\cite{Krizhevsky2012ImageNet}, whereas the last layer acts as a softmax classifier. The activation function for all weight layers (except for the softmax layer) is the rectified linear unit ~\cite{Krizhevsky2012ImageNet}.}
	\resizebox{\textwidth}{10mm}{
		\begin{tabular}{l|@{\hspace{0.5mm}}c@{\hspace{0.5mm}}|@{\hspace{0.5mm}}c@{\hspace{0.5mm}}|@{\hspace{0.5mm}}c@{\hspace{0.5mm}}|@{\hspace{0.5mm}}c@{\hspace{0.5mm}}|@{\hspace{0.5mm}}c@{\hspace{0.5mm}}|@{\hspace{0.5mm}}c
@{\hspace{0.5mm}}|@{\hspace{0.5mm}}c@{\hspace{0.5mm}}|@{\hspace{0.5mm}}c}
			\hline
			%\cline{5-5}
			Arch & conv1 & conv2 & conv3 & conv4 & conv5 & fc6 & fc7 & fc8 \\
			%\cline{5-5}
			\hline
			CNN-F &
			\begin{tabular}[c]{@{}c@{}}$64\times11\times11$\\ st.(4), pad.(0)\\ LRN, x2 pool\end{tabular} & \begin{tabular}[c]{@{}c@{}}$256\times5\times5$\\ st.(1), pad.(2)\\ LRN, x2 pool\end{tabular} & \begin{tabular}[c]{@{}c@{}}$256\times3\times3$\\ st.(1), pad.(1)\\ -\end{tabular} & \begin{tabular}[c]{@{}c@{}}$256\times3\times3$\\ st.(1), pad.(1)\\ -\end{tabular} & \begin{tabular}[c]{@{}c@{}}$256\times3\times3$\\ st.(1), pad.(1)\\ x2 pool\end{tabular} & \begin{tabular}[c]{@{}l@{}}4096\\ drop-\\ out\end{tabular} & \begin{tabular}[c]{@{}l@{}}4096\\ drop-\\ out\end{tabular} & \begin{tabular}[c]{@{}l@{}}1000\\ soft-\\ max\end{tabular} \\ \cline{5-5}
			%\multirow{1}{1cm}{CNN-F} & \multirow{3}{1.8cm}{64x11x11 st. 4, pad 0 LRN, x2 pool} & \multirow{3}{1.8cm}{256x5x5st.1, pad 2 LRN, x2 pool} & %\multirow{3}{1.8cm}{256x3x3st. 1, pad 1 -} & \multirow{3}{1.8cm}{256x3x3\\st. 1, pad 1 - } & \multirow{3}{1.8cm}{256x3x3\\st. 1, pad 1\\ x2 pool} & %\multirow{3}{1cm}{4096 drop-out}& \multirow{3}{1cm}{4096 drop- out}& \multirow{3}{1cm}{1000 soft-max}\\
			\hline	
	\end{tabular}}
	\label{tab:cnn-f}
\end{table}

The proposed multi-view dynamic image adaptive CNN learning model is constructed using the CNN-F model in~\cite{chatfield2014return}, which consists of five convolutional layers and three fully connected layers. The output of the fc6 layer is employed for action representation. The detailed CNN-F model structure is shown in Table~\ref{tab:cnn-f}. During training, most of the training parameters were set as in~\cite{chatfield2014return} with a momentum of 0.9 and a weight decay of 0.0005. However, in the proposed model, the initial learning rate was set to 0.001, i.e., decreased by a factor of 10, and batch size to $8\times4$. The five temporal segments of each depth video illustrated in Sec.~\ref{sec:mvdi_extraction} were randomly chosen for training (i.e., one for each batch). The training procedure was executed on a single NVIDIA GTX 1080 GPU.

As the existing faster R-CNN model~\cite{ren2015faster} is generally trained on RGB object detection datasets, it should be re-trained for human detection in depth video. Thus, sufficient training samples are required. Fortunately, in all three test datasets, human body skeleton information is provided per frame. The minimum bounding rectangle that covers all the skeleton joints is regarded as the ground-truth human bounding box for training. The technical pipeline diagram and the procedure in~\cite{ren2015faster} were strictly followed to train the faster R-CNN. The only difference is that in the present scenario, the input of the faster R-CNN is depth maps instead of RGB images as in~\cite{ren2015faster}.

LIBLINAR was used as the SVM classifier. A linear kernel was applied owing to its efficiency. The penalty factor $C$ was set by 5-fold cross-validation on the training set. After PCA, the dimension of the feature vector for the SVM was reduced to 1000.

The experimental results are organized as follows. The action recognition results for the three benchmark datasets are reported in Secs.~\ref{sec:ntu_rgb-d}--\ref{sec:uwa3dii}. The effectiveness of the dynamic image, multi-view projection, multi-view dynamic image adaptive CNN learning model, spatial-temporal action proposal, and SVM is demonstrated in Secs.~\ref{sec:di-dmm}--\ref{sec:softmax_svm}, respectively.

\begin{figure}[t]
	%h 此处 t 页定 b 页低 p独立一页
	\centering
	\includegraphics[width=9cm]{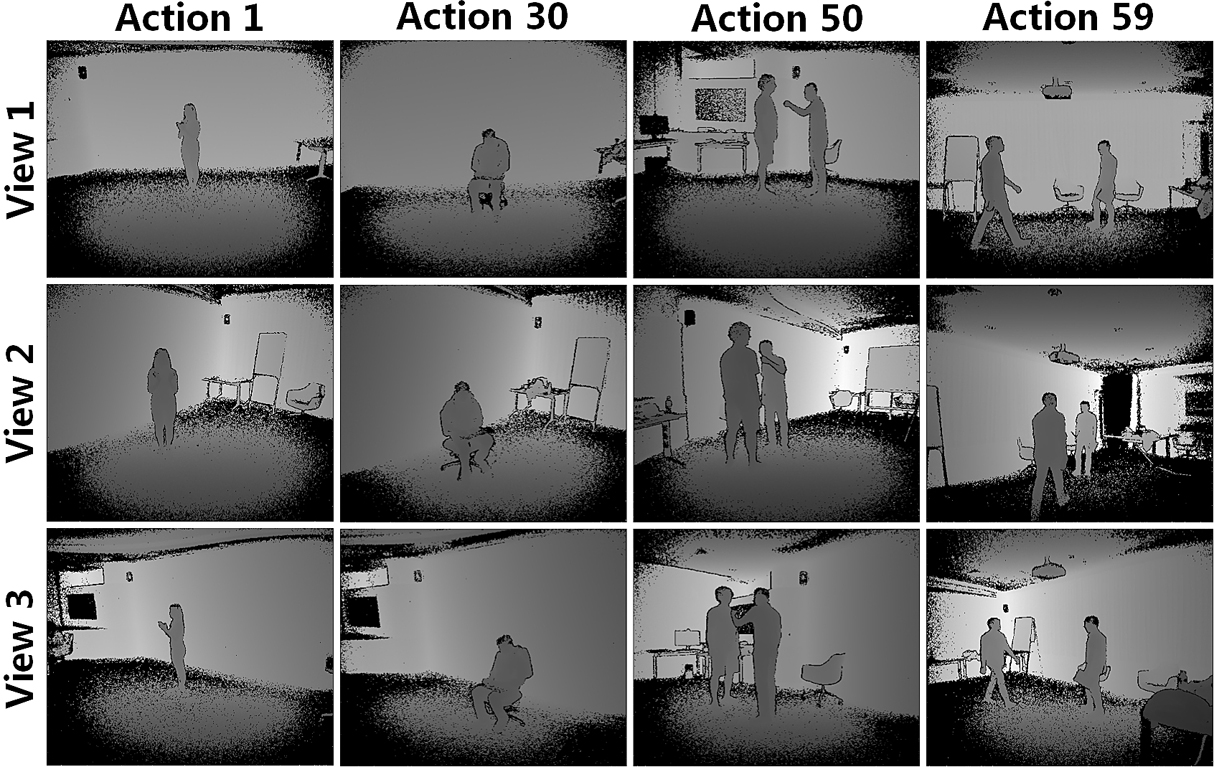}
	\caption{Key depth frames of four actions from three different viewpoints in the NTU RGB-D dataset.}
	\label{fig:ntu_rgb-d}
\end{figure}

\begin{table}[t]
	\linespread{1.05}\selectfont
	\centering
    \footnotesize
	\caption{Comparison of action recognition accuracy (\%) for different methods on the NTU RGB-D dataset.}
	\begin{tabular}{l c c }
		\hline
		\multirow{1}{3cm}{Method} & Cross-subject & Cross-view \\
		\hline
		\multicolumn{3}{c}{\textbf{Input: Skeleton data}}  \\
		%\hline
		Skeletal Quads~\cite{Evangelidis2014Skeletal} & 38.6  & 41.4 \\
		
		LARP~\cite{vemulapalli2014human} & 50.1 & 52.8 \\
		
		HBRNN-L~\cite{Du2015} & 59.1 & 64.0 \\
		
		FTP Dynamic Skeletons~\cite{Hu2015} & 60.2 & 65.2 \\
		
		PA-LSTM~\cite{Shahroudy_2016_CVPR} & 63.0 & 70.3 \\
		
		ST-LSTM~\cite{Liu2016Spatio} & 69.2 & 77.7\\

        GCA-LSTM network~\cite{liu2017} & 74.4 & 82.8\\

        Clips+CNN+MTLN~\cite{ke2017} & 79.6 & 84.8\\
		
		\hline
		\multicolumn{3}{c}{\textbf{Input: Depth maps}}  \\
		%\hline
		
		HON4D~\cite{oreifej2013hon4d} & 30.6 & 7.3 \\
		
		SNV~\cite{yang2014super} & 31.8 & 13.6 \\
		
		HOG$^{2} $~\cite{Ohn2013} & 32.2 & 22.3 \\
		
		%\hline
	%	Ours(MBB-MVMTDI+softmax) & 71.8 & 73.7\\
	%	Ours(O-MVMTDI+svm) & 78.6 & 75.4\\
		Multi-view dynamic images+CNN & \textbf{84.6} & \textbf{87.3} \\
		
		\hline
		
	\end{tabular}
	\label{tab:ntu-rgbd}
\end{table}

\subsection{NTU RGB+D dataset}~\label{sec:ntu_rgb-d}
NTU RGB+D is a recently proposed large-scale dataset for RGB-D human action recognition. In particular, it involves 56000 samples of 60 action classes, collected from 40 subjects. The actions can be generally divided into three categories: 40 daily actions (e.g., \emph{drinking, eating, reading}), nine health-related actions (e.g., \emph{sneezing, staggering, falling down}), and 11 mutual actions (e.g., \emph{punching, kicking, hugging}). These actions take place under 17 different scene conditions corresponding to 17 video sequences (i.e., S001--S017). The actions were captured using three cameras with different horizontal imaging viewpoints, namely, $-45^{\circ}, 0^{\circ}$, and $+45^{\circ}$. Fig.~\ref{fig:ntu_rgb-d} shows some key depth frames of four actions in this dataset from three different viewpoints. Multi-modality information is provided for action characterization, including depth maps, 3D skeleton joint position, RGB frames, and infrared sequences. The performance evaluation was performed by a cross-subject test that split the 40 subjects into training and test groups, and by a cross-view test that employed one camera ($+45^{\circ}$) for testing, and the other two cameras for training. The results of the comparison between the proposed method and other state-of-the-art approaches are listed in Table~\ref{tab:ntu-rgbd}. The following can be observed:

$\bullet$ The proposed action recognition method consistently outperforms all the other state-of-the-art approaches in both the cross-subject and cross-view tests, regardless of whether the input was skeleton data or depth maps. This demonstrates the effectiveness of the proposed method for depth-based action recognition under complex scene and viewpoint conditions.

$\bullet$ The proposed method significantly outperforms the other depth-map-based approaches by large margins. This may be due to the following: (1) The hand-crafted visual descriptors employed by the other approaches are not as discriminative as the proposed CNN-based feature learning in multi-view dynamic images. (2) The other methods are sensitive to viewpoint variation. However, the proposed multi-view projection in depth video can alleviate this. It should be noted that the proposed approach can even achieve better performance in the cross-view test. (3) The other methods take the entire depth maps into consideration for action characterization without effective action proposal. Thus, they are also sensitive to scene and action position variation.

$\bullet$ Skeleton-based approaches achieve considerably better performance than depth-map-based approaches~\cite{oreifej2013hon4d,yang2014super,Ohn2013}, except for the proposed method. In fact, the most recently proposed depth-based action recognition techniques~\cite{ke2017,liu2017} mainly resort to skeleton information. However, the present study verifies that  using the depth maps only can also achieve comparable or even better performance by employing feature learning and human detection.

\subsection{Northwestern-UCLA dataset}~\label{sec:northwestern-ucla}

\begin{figure}[t]
	%h 此处 t 页定 b 页低 p独立一页
	\centering
	\includegraphics[width=9cm]{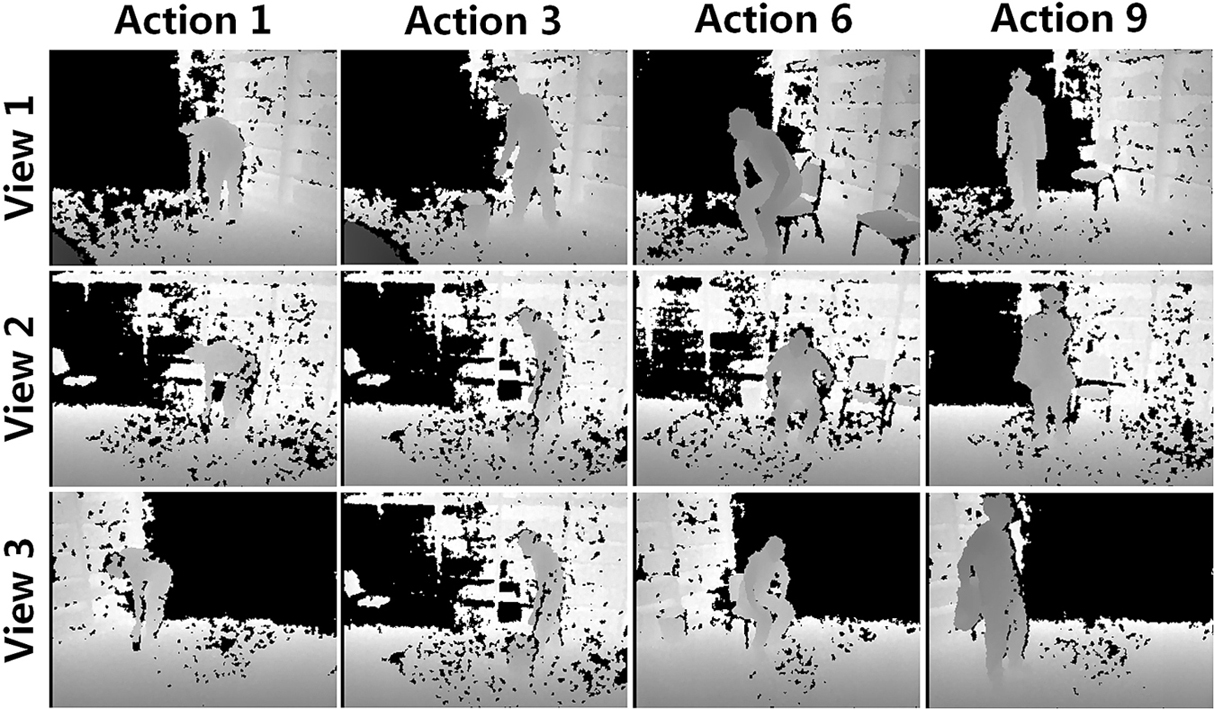}
	\caption{Key depth frames of four actions from three different viewpoints in the Northwestern--UCLA dataset.}
	\label{fig:northwestern-ucla}
\end{figure}

\begin{table}[t]
	\linespread{1.05}\selectfont
	\centering
    \footnotesize
	\caption{Comparison of action recognition accuracy (\%) for different methods on the Northwestern--UCLA dataset.}
	\begin{tabular}{lc}
		\hline
		Method & Accuracy\\
		\hline
		\multicolumn{2}{c}{\textbf{Input: Skeleton data}}  \\
		%\hline
		HOJ3D \cite{xia2012view} & 54.4 \\
		
		Actionlet \cite{wang2014learning}  & 69.9 \\

		LARP \cite{vemulapalli2014human} & 74.2  \\
		
		\hline
		\multicolumn{2}{c}{\textbf{Input: Depth maps+human mask}}  \\
		%\hline
		HPM+TM+external training data~\cite{Rahmani} & \textit{92.0} \\
		
		\hline
		\multicolumn{2}{c}{\textbf{Input: Depth maps}}  \\
		%\hline
		
		CCD~\cite{Cheng2012Human} & 34.4 \\
		HON4D~\cite{oreifej2013hon4d} & 39.9 \\
		SNV~\cite{yang2014super} & 42.8 \\
		DVV~\cite{Li2012Discriminative} &  52.1 \\
		AOG~\cite{Wang2014} & 53.6 \\
		
		HOPC~\cite{rahmani2014hopc} & 80.0 \\
		
		%\hline
		Multi-view dynamic images+CNN & \textbf{84.2}  \\
		
		\hline
		
	\end{tabular}
	\label{table:northwestern-ucla}
\end{table}

This dataset contains 1475 video samples of 10 action categories: \emph{picking up with one hand, picking up with two hands, dropping off trash, walking around, sitting down, standing up, donning, doffing, throwing, and carrying.} They were captured by three depth cameras from different viewpoints in varying scene conditions. Each action was executed by 10 subjects. Fig.~\ref{fig:northwestern-ucla} shows some key depth frames of four actions in this dataset from three different viewpoints. Compared with NTU RGB-D, the samples in this dataset contain considerably higher imaging noise, which makes action recognition significantly more difficult. Multi-modality information is provided for action characterization, including depth maps, 3D skeleton joint position, and RGB frames. Following~\cite{wang2014cross}, the samples from the first two cameras were used for training, and the samples from the third camera for testing. The results of the comparison between the proposed approach and the other state-of-the-art approaches are listed in Table~\ref{table:northwestern-ucla}. The following can be noted:

$\bullet$ Except for HPM+TM~\cite{Rahmani}, the proposed approach can still achieve better performance than the other state-of-the-art methods when the input is skeleton data or depth maps. This verifies the superiority and generality of the proposed method on different datasets.

$\bullet$ The recognition accuracy of HPM+TM ~\cite{Rahmani} is higher than that of the proposed method. However, it introduces external synthetic data to train the CNN. Moreover, accurate human masks are also required to counter the effect of background and imaging noise. By contrast, the proposed approach does not require these. Thus, without external training data, the proposed method outperforms all the others.

$\bullet$ The training sample size for this dataset is significantly smaller than that for the NTU RGB+D dataset. Nevertheless, the proposed approach can still achieve good performance (i.e., 84.2\%). This demonstrates that the proposed action recognition method is applicable to both small-scale and large-scale cases.

\begin{figure}[t]
	%h 此处 t 页定 b 页低 p独立一页
	\centering
	\includegraphics[width=9cm]{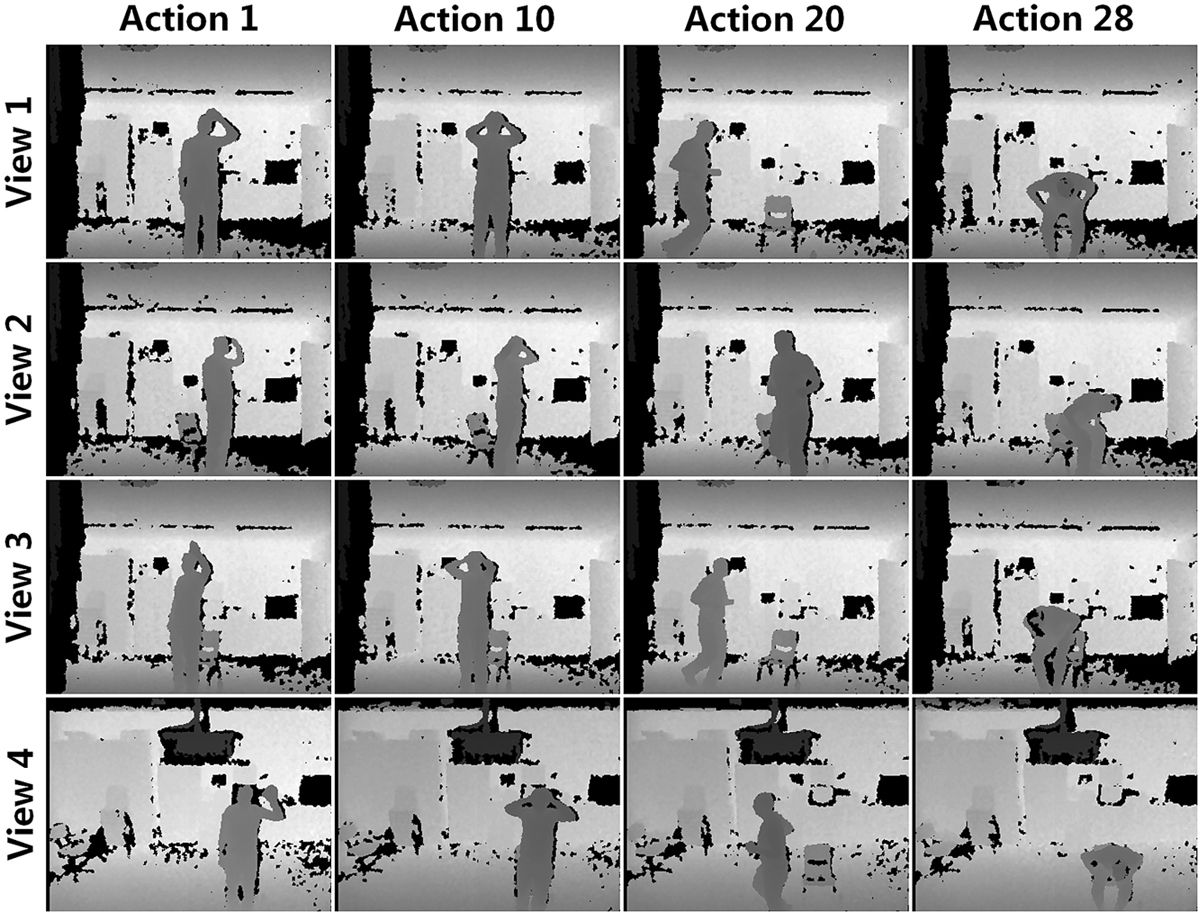}
	
	\caption{Key depth frames of four actions from four different viewpoints in the UWA3DII dataset.}
	\label{fig:uwa3d}
\end{figure}

\begin{table*}[t]
	\linespread{1.05}\selectfont
	\centering
    \scriptsize
	\caption{Comparison of action recognition accuracy (\%) for different methods on the UWA3DII dataset. Each time, two views are used for training, and the remaining two views are for testing. In particular, ``MVDI" indicates the proposed multi-view dynamic images, and ``HPM+TM" runs with the external training data.}
	\begin{tabular}{@{\hspace{1mm}}l@{\hspace{1mm}}@{\hspace{1mm}}c@{\hspace{1mm}}@{\hspace{1mm}}c@{\hspace{1mm}}@{\hspace{1mm}}c@{\hspace{1mm}}@{\hspace{1mm}}c@{\hspace{1mm}}@{\hspace{1mm}}c@{\hspace{1mm}}
@{\hspace{1mm}}c@{\hspace{1mm}}@{\hspace{1mm}}c@{\hspace{1mm}}@{\hspace{1mm}}c@{\hspace{1mm}}@{\hspace{1mm}}c@{\hspace{1mm}}@{\hspace{1mm}}c@{\hspace{1mm}}@{\hspace{1mm}}c@{\hspace{1mm}}
@{\hspace{1mm}}c@{\hspace{1mm}}@{\hspace{1.5mm}}c@{\hspace{1.5mm}}}
		\hline
		Training views & \multicolumn{2}{c}{$V_1\&V_2$} & \multicolumn{2}{c}{$V_1\&V_3$} & \multicolumn{2}{c}{$V_1\&V_4$} & \multicolumn{2}{c}{$V_2\&V_3$} & \multicolumn{2}{c}{$V_2\&V_4$} & \multicolumn{2}{c}{$V_3\&V_4$} & \multirow{2}{0.6cm}{Mean}\\
		
		%\cline{1-13}
		Test views &  $V_3$ & $V_4$ & $V_2$ & $V_4$ & $V_2$ & $V_3$ & $V_1$ & $V_4$ & $V_1$ & $V_3$ & $V_1$ & $V_2$ & \\
		\hline
		\multicolumn{14}{c}{\textbf{Input: Skeleton data}}\\
		%\hline
		
		HOJ3D \cite{xia2012view} & 15.3 & 28.2 & 17.3 & 27.0 & 14.6 & 13.4 & 15.0 & 12.9 & 22.1 & 13.5 & 20.3 & 12.7 & 17.7 \\
		
		Actionlet \cite{wang2014learning} & 45.0 & 40.4 & 35.1 & 36.9 & 34.7 & 36.0 & 49.5 & 29.3 & 57.1 & 35.4 & 49.0 & 29.3 & 39.8 \\
		
		LARP \cite{vemulapalli2014human} & 49.4 & 42.8 & 34.6 & 39.7 & 38.1  & 44.8 & 53.3 & 33.5 & 53.6 & 41.2 & 56.7 & 32.6  & 43.4 \\
		\hline
		\multicolumn{14}{c}{\textbf{Input: Depth maps+human mask}}\\
		%\hline
		
		HPM+TM~\cite{Rahmani} & \textit{80.6} & \textit{80.5} & \textit{75.2} & \textit{82.0} & \textit{65.4} & \textit{72.0} & \textit{77.3} & \textit{67.0} & \textit{83.6} &\textit{ 81.0} & \textit{83.6} & \textit{74.1} & \textit{76.9} \\
	
		\hline
		\multicolumn{14}{c}{\textbf{Input: Depth maps}}\\
		%\hline
		
		CCD~\cite{Cheng2012Human} & 10.5 & 13.6 & 10.3 & 12.8 & 11.1 & 8.3 & 10.0 & 7.7 & 13.1 & 13.0 & 12.9 & 10.8 & 11.2 \\
		DVV~\cite{Li2012Discriminative} & 23.5 & 25.9 & 23.6 & 26.9 & 22.3 & 20.2 & 22.1 & 24.5 & 24.9 & 23.1 & 28.3 & 23.8 & 24.2 \\
		AOG~\cite{Wang2014} & 23.9 & 31.1 & 25.3 & 29.9 & 22.7 & 21.9 & 25.0 & 20.2 & 30.5 & 27.9 & 30.0 & 26.8 & 26.7 \\
		HON4D~\cite{oreifej2013hon4d} & 31.1 & 23.0 & 21.9 & 10.0 & 36.6 & 32.6 & 47.0 & 22.7 & 36.6 & 16.5 & 41.4 & 26.8 & 28.9 \\
		SNV~\cite{yang2014super} & 31.9 & 25.7 & 23.0 & 13.1 & 38.4 & 34.0 & 43.3 & 24.2 & 36.9 & 20.3 & 38.6 & 29.0 & 29.9 \\
		HOPC~\cite{rahmani2014hopc} & 52.7 & 51.8 & 59.0 & \textbf{57.5} & 42.8 & 44.2 & 58.1 & 38.4 & 63.2 & 43.8 & 66.3 & 48.0 & 52.2 \\
		%\hline
		
		MVDI+CNN & \textbf{77.0} & \textbf{59.5} & \textbf{68.3} & 57.2 & \textbf{57.8} & \textbf{72.9 }& \textbf{80.3} & \textbf{51.3} & \textbf{76.6} & \textbf{69.5} & \textbf{78.8} & \textbf{67.9} & \textbf{68.1} \\
		
		\hline
		
	\end{tabular}
	\label{tab:uwa3d}
\end{table*}

\subsection{UWA3DII dataset}~\label{sec:uwa3dii}
UWA3DII consists of 1075 action samples from 30 classes: \emph{one hand waving, one hand punching, two hands waving, two hands punching, sitting down, standing up, vibrating, falling down, holding chest, holding head, holding back, walking, irregular walking, lying down, turning around, drinking, phone answering, bending, jumping jack, running, picking up, putting down, kicking, jumping, dancing, moping floor, sneezing, sitting down (chair), squatting, and coughing.} These actions were captured using 10 subjects from four different viewpoints (i.e., front, top, left, and right) under the same scene conditions. Multi-modality information was provided for action characterization, including depth maps, 3D skeleton joint position, and RGB frames. This dataset is challenging because the action samples were acquired from varying viewpoints at different time points. Moreover, serious self-occlusion may occur. Finally, some action categories are of high similarity. Fig.~\ref{fig:uwa3d} shows some key depth frames of four actions in this dataset from four different viewpoints. Following~\cite{rahmani2016histogram}, the samples from two viewpoints were used for training, and the remaining two for testing. Cross-validation among the viewpoints was performed. The performance comparison between the proposed approach and the other state-of-the-art approaches is listed in Table~\ref{tab:uwa3d}. The following can be observed:

$\bullet$ Regarding the different viewpoint combinations, the proposed method significantly outperforms all the other approaches nearly in all cases, except for HPOC~\cite{rahmani2014hopc} (i.e., training on $V_1\&V_3$ and testing on $V_4$) and HPM+TM~\cite{Rahmani}. As previously mentioned, HPM+TM employs external training data and human masks. Without these, the proposed approach achieves the best overall performance among all the methods under comparison (i.e., $15.9\%$ better than the second best). This demonstrates the superiority of the proposed method for depth-based action recognition under varying viewpoint conditions.

$\bullet$ The performance of the proposed approach on this dataset (i.e., overall $68.1\%$) is relatively poor compared with that on the NTU RGB-D and Northwestern--UCLA datasets. This may be because the training sample size for this dataset is not sufficient for CNN training. Moreover, the actions from different categories (e.g., drinking vs. phone answering) are of high similarity. It seems that dynamic images cannot capture and emphasize fine-grained action characterization clues very well, which should be addressed in future studies.

$\bullet$ The performance of skeleton-based methods is indeed poor (i.e., $43.4\%$ at most) on this dataset. This verifies that under dramatic viewpoint variation and serious self-occlusion conditions, accurate action representation using skeleton information is still a challenging task. As demonstrated in this study, using depth maps directly can alleviate this to some degree.

\subsection{Comparison between dynamic images and DMMs}~\label{sec:di-dmm}
As previously mentioned, DMMs~\cite{yang2012recognizing} are similar to dynamic images, as they can also summarize the depth video into a single image as
\begin{equation}
\label{eq:dmm}
DMM = \sum_{i=1}^{N-1}(|I_{i+1}-I_i|>\epsilon),
\end{equation}
where $I_i$ denotes the \emph{i}-th depth frame, $|I_{i+1}-I_i|>\epsilon$ indicates the binary motion energy map, and $\epsilon$ is a predefined threshold~\footnote{$\epsilon$ was set following ~\cite{chen2015action}.}. An intuitive comparison of a dynamic image and a DMM is shown in Fig.~\ref{fig:mvdi_dmm}. In particular, they correspond to the same ``\emph{Hugging other person}" action in the NTU RGB-D dataset. The following can be observed:

\begin{figure}[t]
	\centering

    \subfigure[Dynamic image]{\includegraphics[width=4.4cm]{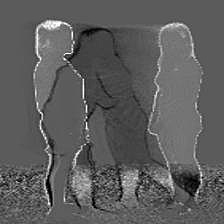}}
    \subfigure[DMM]{\includegraphics[width=4.4cm]{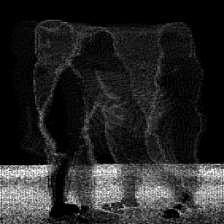}}
	
	\caption{Dynamic image and DMM for ``\emph{Hugging other person}" in the NTU RGB-D dataset.}
	
\label{fig:mvdi_dmm}
\end{figure}

\begin{table}[t]
	\centering
    %\linespread{1.05}\selectfont
    \footnotesize
	\caption{Comparison of action recognition accuracy (\%) between dynamic image and DMM on the three test datasets.}
    \begin{tabular}{l c c c c}
		\hline
		\multicolumn{3}{c}{Dataset / Test setting} & DMM & Dynamic image \\
		
		\hline
		
		\multirow{1}{1.55cm}{NTU RGB-D} & \multicolumn{2}{|c}{Cross-subject} & 74.7 & \textbf{84.6} \\
		%\cline{2-5}
		& \multicolumn{2}{|c}{Cross-view} & 72.1 & \textbf{87.3}\\		
		\hline
		\multicolumn{3}{l}{Northwestern-UCLA} & 78.4 & \textbf{84.2} \\
		
		%\hline

        \multicolumn{3}{l}{UWA3D II} & 51.5 & \textbf{68.1} \\
		
		\hline
	\end{tabular}

	\label{tab:di-dmm}
\end{table}

$\bullet$ Dynamic images can reveal the motion temporal order of actions by using the gray-scale value, which cannot be achieved by a DMM.

$\bullet$ Compared with DMMs, dynamic images better suppress the effect of background and imaging noise.

$\bullet$ Owing to the unsuitable setting of the motion threshold (i.e., $\epsilon$ in Eqn.~\ref{eq:dmm}), DMMs tend to lose some action details of relatively low motion energy. This can be detrimental for effective action representation.

A quantitative performance comparison between dynamic images and DMMs is now performed on all three test datasets. They share the same multi-view projection procedure, CNN learning model, and action proposal results. The results of the comparison are listed in Table~\ref{tab:di-dmm}. It can be seen that dynamic imaging significantly outperforms DMMs in all test cases. It verifies the superiority of dynamic imaging in depth video summarization for action characterization.

\begin{table}[t]
	\linespread{1.05}\selectfont
    \footnotesize
	\centering
	\caption{Comparison of action recognition accuracy (\%) among viewpoint groups in Table~\ref{tab:view-group} and their combination in the S001 sequence of the NTU RGB-D dataset.}

    \begin{tabular}{l | c  c  c}
		\hline
		%\multirow{3}{3cm}{Method} & Cross-subject & Cross-view\\
        View group & Cross-subject & Cross-view\\		
		\hline
		 Group 1  &  76.7  & 75.1 \\
         Group 2 &  78.0  &  88.4 \\
         Group 3 &  70.4  &  82.8 \\
         Group 4 &  79.8  &  91.4 \\
         Group 5 &  78.6  &  76.5 \\
        \hline
        Group 1+2  &  80.2  &  89.5 \\
        Group 1+2+3  &  81.6  &  93.1 \\
        Group 1+2+3+4  &  84.0  &  94.3 \\
        Group 1+2+3+4+5  &  \textbf{84.1}  & \textbf{94.9} \\		
        \hline

	\end{tabular}

	\label{tab:mvp_effect}
\end{table}

\subsection{Effectiveness of multi-view projection}~\label{sec:mvp_effect}
To verify the effectiveness of multi-view projection in depth video, the performance of the view groups in Table~\ref{tab:view-group} and their combination are compared using the proposed action recognition method. The test was executed only on the S001 action sequence of the NTU RGB-D dataset, owing to the high computational cost. The results of the comparison are listed in Table~\ref{tab:mvp_effect}. The following can be noted:

$\bullet$ As view groups increase, the performance of action recognition is consistently enhanced both in the cross-subject and the cross-view tests. It is noteworthy that the performance improvement in the cross-subject test demonstrates that multi-view projection in depth video not only alleviates the cross-view divergence problem but also introduces richer discriminative information for cross-subject action characterization. The results demonstrate the effectiveness of the multi-view projection mechanism in depth video for action recognition.

$\bullet$ As listed in Table~\ref{tab:view-group}, group 3 corresponds to the raw depth view. However, it is interesting that among all the view groups, it does not achieve the best performance. First, this indeed demonstrates the effectiveness of the view projection method proposed in Sec~\ref{sec:mvp_dv}. Second, it reveals the fact that the obtained multi-view projected depth videos do not merely play the role of providing additional auxiliary information.

\subsection{Effectiveness of multi-view dynamic image adaptive CNN learning model}~\label{sec:cnn_effect}

\begin{table}[t]
	\centering
    %\linespread{1.05}\selectfont
    \footnotesize
	\caption{Comparison of action recognition accuracy (\%) among different CNN learning models on all test datasets.}
	\begin{tabular}{l | c c c c}
		\hline
		Dataset & OS CNN & MS CNN & Our CNN\\
		\hline
		NTU RGB-D  &  78.8 & 82.5 &\textbf{84.6}\\
		Northwestern-UCLA   & 82.2 & 82.7  &  \textbf{84.2}\\
		UWA3DII  & 57.1 & 66.4  &  \textbf{68.3}  \\
		
		\hline
	\end{tabular}
	\label{table:cnn_effect}
\end{table}

To demonstrate the superiority of the proposed multi-view dynamic image adaptive CNN learning model, it was compared with two related models. One is the multi-stream CNN learning (MS CNN)  in~\cite{wang2015convnets,wang2016action}. That is, the view groups correspond to the individual convolutional and fully connected layers. The other is the standard one-stream CNN learning (OS CNN)  model~\cite{Jia2014Caffe}. In particular, the view groups were regarded as different input channels. The comparison was performed on the three testing datasets with the same experimental settings. That is, PCA and SVM were also applied to MS CNN and OS CNN. In the NTU RGB-D dataset, the result on cross-subject test is reported. The comparison results are listed in Table~\ref{table:cnn_effect}. The following can be observed:

$\bullet$ In all test datasets, the proposed multi-view dynamic image adaptive CNN learning model consistently outperforms the other CNN models. This demonstrates the effectiveness and generality of the proposed method with multi-view dynamic images for action representation.

$\bullet$ Among the CNN models, the one-stream model is the weakest because its structure of only one-stream fc layers is not suitable for capturing the complementary information from different virtual imaging viewpoints.

\subsection{Effectiveness of spatial-temporal action proposal}~\label{sec:app_effect}

\begin{table}[t]
	%\linespread{1.05}\selectfont
    \footnotesize
	\centering
	\caption{Comparison of action recognition accuracy (\%) of the proposed action recognition method with and without spatial-temporal action proposal on the three test datasets.}
	\begin{tabular}{l| c c c c }
		\hline
		\multicolumn{3}{c}{Dataset / Test setting} & MVDI-O  & MVDI-AP \\
		\hline
		
		\multirow{1}{1.55cm}{NTU RGB-D} & \multicolumn{2}{c}{Cross-subject} & 78.6 & \textbf{84.6} \\
		%\cline{2-5}
		& \multicolumn{2}{c}{Cross-view} & 75.4  & \textbf{87.3}\\		
		\hline
		\multicolumn{3}{l}{Northwestern-UCLA} &  73.2  & \textbf{84.2} \\
		%\hline

        \multicolumn{3}{l}{UWA3D II} &  \textbf{72.6}  & 68.1 \\
		
		\hline
	\end{tabular}
	\label{tab:ap_effect}
\end{table}

\begin{figure}[t]
	\centering

    \subfigure[Human--human interaction]{\includegraphics[width=4.8cm]{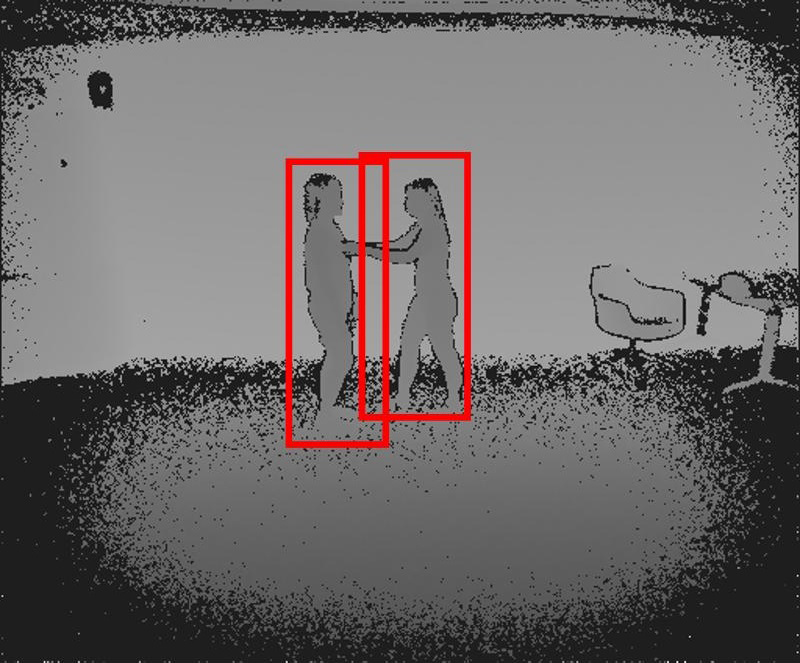}}
    \subfigure[Human--object interaction]{\includegraphics[width=4.8cm]{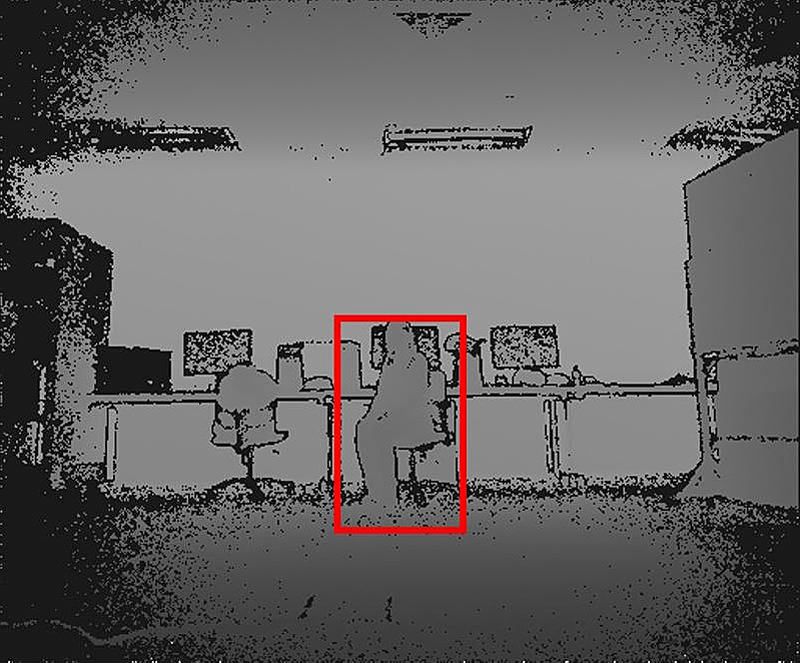}}
	
	\caption{Human detection results of faster R-CNN when human--human or human--object interaction occurs.}
	
\label{fig:rcnn_result}
\end{figure}

In the proposed method, spatial-temporal action proposal is performed to counter the effect of scene variation and spatial-sensitivity of the CNN. To demonstrate its effectiveness, the test cases with and without action proposal were compared in the three test datasets. The results of the comparison are listed in Table~\ref{tab:ap_effect}, where the proposed action recognition method with and without action proposal is denoted by ``MVDI-AP" and ``MVDI-O", respectively. The following can be observed:

$\bullet$ Action proposal can significantly improve the performance of the  method on the NTU RGB-D and Northwestern--UCLA datasets in three cases. The performance enhancement is $6.0\%$ at least. This verifies the feasibility and effectiveness of the spatial-temporal action proposal approach. It also demonstrates that countering the effect of scene variation and spatial sensitivity of the CNN is essential for action recognition.

$\bullet$ In the UWA3D II dataset, when action proposal is employed, the performance of the proposed approach drops. This may be because the action samples in this dataset are captured under similar scene conditions. Thus, in this case, performing action proposal tends to cause information loss. The enhancement of the adaptability of action proposal will be addressed in future studies.

Furthermore, human detection using faster R-CNNs plays a key role for action proposal. It was demonstrated that faster R-CNNs is applicable to depth frames even when human--human or human--object interaction occurs. Fig.~\ref{fig:rcnn_result} shows some live examples.

\subsection{Comparison between softmax classifier and SVM}~\label{sec:softmax_svm}

\begin{table}[t]
	%\linespread{1.05}\selectfont
    \footnotesize
	\centering
	\caption{Comparison of action recognition accuracy (\%) of softmax classifier and SVM on the three test datasets.}
	\begin{tabular}{l |c c c c }
		\hline
		\multicolumn{3}{c}{Dataset / Test setting} & Softmax  & SVM \\
		\hline
		
		\multirow{1}{1.55cm}{NTU RGB-D} & \multicolumn{2}{c}{Cross-subject} & 71.8 & \textbf{84.6} \\
		%\cline{2-5}
		& \multicolumn{2}{c}{Cross-view} & 73.7 & \textbf{87.3}\\		
		\hline

		\multicolumn{3}{l}{Northwestern-UCLA} &  70.0  & \textbf{84.2} \\
		%\hline

        \multicolumn{3}{l}{UWA3D II} &  57.0  & \textbf{68.1} \\
		
		\hline
	\end{tabular}
	\label{tab:softmax_svm}
\end{table}

In the proposed action recognition approach, an SVM is employed as the classifier for deciding the action category instead of the softmax classifier in the CNN. The intuition is that the training sample size of depth actions is still insufficient to tune the CNN well in end-to-end learning. The introduction of the SVM can alleviate this. To verify the superiority of the SVM, it was compared with the softmax classifier on the three test datasets with the same experimental settings and CNN structure. It is noteworthy that in the proposed CNN model, multi-stream softmax classifiers corresponding to different view groups are involved. In softmax-based classification, for a test action sample $X$, the output of different softmax classifiers is combined by summation to acquire the final classification score. The results of the comparison between the softmax classifier and the SVM are listed in Table~\ref{tab:softmax_svm}. It can be summarized that in all cases, the SVM is significantly better than the softmax classifier (i.e., $11.1\%$ better at least). This demonstrates that for a specific depth-based action recognition task, intuitively applying CNNs with end-to-end learning but without sufficient training samples is not the optimal choice. Alleviating this by unsupervised or low-shot learning will be addressed in future studies.

\begin{table}[t]
	\centering
	%\linespread{1.05}\selectfont
	\footnotesize
	\caption{Average online time consumption per video corresponding to the main technical components in the proposed approach.}
	\begin{tabular}{l | c }
		\hline
		Technical components & Time cost (s)\\
		\hline
		Multi-view projection (CPU) & 34.5 \\
		Multi-view dynamic images extraction (CPU)  &  5.6 \\
		Spatial-temporal action proposal (GPU)  & 2.4 \\
		CNN feature extraction (GPU) & 8.5 \\
		SVM classification (CPU) & 0.02\\
		\hline
		Overall time cost & 51.02\\
		\hline
	\end{tabular}
	\label{table:time_cost}
\end{table}

\subsection{Time consumption of proposed approach}~\label{sec:time_consume}
Herein, the online running efficiency of the proposed method will be analyzed. 100 video samples were randomly selected from the NTU RGB-D dataset for testing, with an average length of 97 frames per video sample. The time consumption (excluding I/O time) statistics were computed on an Intel (R) Xeon(R) E5-2630 V3 computer running at 2.4 GHz (using only one core) with an Nvidia GeForce 1080 GPU. The average time consumption per video corresponding to the main technical components is listed in Table~\ref{table:time_cost}. It can be observed that multi-view projection is the most time consuming. The running efficiency of the proposed approach is not high. Its acceleration is a critical issue that should be addressed in future work for practical applications.

\section{Conclusions}~\label{sec:conclusion}
A novel action recognition approach based on dynamic image for depth video was proposed. Through multi-view projection in depth video, multi-view dynamic images are extracted for action characterization by summarizing the depth video into static images. The key insight for this is to involve more discriminative information concerning the 3D features of depth video. To better handle multi-view dynamic images, a novel CNN learning model was also proposed. That is, different view groups share the same convolutional layers but with different fully connected layers. The main advantage of the proposed CNN model is the alleviation of the gradient vanishing problem of CNN training, particularly on the shallow convolutional layers. Spatial-temporal action proposal is performed to counter the effect of scene variation and spatial-sensitivity of the CNN. The experimental results on three test datasets demonstrated the superiority of the proposed approach.

In future work, it is planned to embed multi-view dynamic images into deeper CNN structure (i.e., ResNet~\cite{he2016deep}) by relaxing the strong requirement on training sample size. The enhancement of the generality of the proposed spatial-temporal action proposal approach for different scene conditions will also be considered. Another direction is to resort to unsupervised deep learning as well as weakly and semi-supervised learning to alleviate the problem of  insufficient labeled depth action video samples for training.

\section{Acknowledgements}
This work is jointly supported by the Natural Science Foundation of China (Grant No. 61502187, 61876211, 61602193 and  61702182), the National Key R\&D Program of China (No. 2018YFB1004600), the International Science \& Technology Cooperation Program of Hubei Province, China (Grant No. 2017AHB051), the HUST Interdisciplinary Innovation Team Foundation (Grant No. 2016JCTD120), and Hunan Provincial Natural Science Foundation of China (Grant 2018JJ3254). Joey Tianyi Zhou is supported by Programmatic Grant No. A1687b0033 from the Singapore government's Research, Innovation and Enterprise 2020 plan (Advanced Manufacturing and Engineering domain).

\bibliographystyle{elsarticle-harv}
\bibliography{reference}

\end{document}